\documentclass{article} % For LaTeX2e
\usepackage{collas2025_conference,times}
\usepackage{easyReview}
\usepackage{wrapfig}
\usepackage{placeins}
\usepackage{booktabs}
\collasfinalcopy

\usepackage{amsmath,amsfonts,bm}

\def\eqref#1{equation~\ref{#1}}
\def\1{\bm{1}}

\DeclareMathAlphabet{\mathsfit}{\encodingdefault}{\sfdefault}{m}{sl}
\SetMathAlphabet{\mathsfit}{bold}{\encodingdefault}{\sfdefault}{bx}{n}

\usepackage{hyperref}
\usepackage{url}
\usepackage{xcolor}

\title{Balancing Expressivity and Robustness: Constrained Rational Activations for Reinforcement Learning}

\author{Rafał Surdej \\
Warsaw University of Technology, Poland \\
\texttt{rafalsurdej@gmail.com}\\
\And Michał Bortkiewicz \\
Warsaw University of Technology, Poland \\
\And Alex Lewandowski \\
University of Alberta, Edmonton, Canada
\And Mateusz Ostaszewski\thanks{Equal supervision.} \\
Warsaw University of Technology, Poland \\
\And Clare Lyle$^*$ \\
Google DeepMind}

\begin{document}

\maketitle

\begin{abstract}

 %Trainable activation functions, whose parameters are optimized alongside network weights, offer increased expressivity compared to fixed activation functions.
 %Specifically, rational activation functions have been proposed as a way to enhance plasticity in reinforcement learning.
 %However, their impact on training stability remains unclear.
%We investigate trainable rational activations in challenging continuous control tasks and find that the same properties that enhance adaptability can also introduce instability due to overestimation errors. To address this, we introduce structural constraints that prevent excessive activation scaling while preserving adaptability. Experiments across MetaWorld and DeepMind Control Suite (DMC) environments show that our approach significantly improves training stability and overall performance, outperforming both standard rational activations and ReLU-based baselines.
%Our findings demonstrate the effectiveness of structured activation constraints in RL and provide new insights into activation function design for improving robustness in dynamic learning environments. The source code of our experiments is available at: \url{https://anonymous.4open.science/r/rl_plasticity-4BAF} 

Trainable activation functions, whose parameters are optimized alongside network weights, offer increased expressivity compared to fixed activation functions. Specifically, trainable activation functions defined as ratios of polynomials (rational functions) have been proposed to enhance plasticity in reinforcement learning. However, their impact on training stability remains unclear.
In this work, we study trainable rational activations in both reinforcement and continual learning settings. We find that while their flexibility enhances adaptability, it can also introduce instability, leading to overestimation in RL and feature collapse in longer continual learning scenarios. Our main result is demonstrating a trade-off between expressivity and plasticity in rational activations. To address this, we propose a constrained variant that structurally limits excessive output scaling while preserving adaptability. Experiments across MetaWorld and DeepMind Control Suite (DMC) environments show that our approach improves training stability and performance. In continual learning benchmarks, including MNIST with reshuffled labels and Split CIFAR-100, we reveal how different constraints affect the balance between expressivity and long-term retention.
While preliminary experiments in discrete action domains (e.g., Atari) did not show similar instability, this suggests that the trade-off is particularly relevant for continuous control. Together, our findings provide actionable design principles for robust and adaptable trainable activations in dynamic, non-stationary environments.
Code available at: \url{https://github.com/special114/rl_rational_plasticity}.
\end{abstract}

\section{Introduction}

\begin{wrapfigure}[19]{r}{0.5\linewidth}%[<liczba linii>]{<pozycja>}{<szerokość>}
    \vspace{-15pt}
    \centering
    \includegraphics[width=\linewidth]{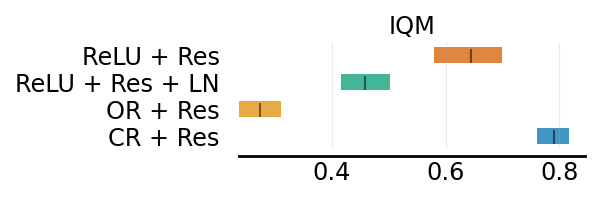}
    \caption{Interquartile Mean (IQM) performance after 1M environment steps, aggregated across 15 MetaWorld and 15 DeepMind Control Suite (DMC) environments. For MetaWorld, we measure the score, while for DMC, returns are divided by 1000 to match the upper performance bound. We compare Original Rationals (OR), our Constrained Rationals (CR), ReLU, and ReLU with Layer Normalization (LN), all trained with resets. Our results show that CR + Resets achieves the highest overall performance, highlighting the benefits of our proposed constraints in stabilizing RL training.}
    \label{fig:dmc-mw-combined-subset}
\end{wrapfigure}

Neural network expressivity is a key factor in reinforcement learning (RL), particularly in dynamic environments where agents must continuously adapt. While most RL architectures rely on static activation functions, recent work suggests that allowing activations to be trainable could enhance adaptability by increasing the flexibility of individual neurons. By co-evolving with learned features, trainable activations can dynamically adjust their shape, improving expressivity beyond fixed functions. In particular, rational activation functions~\citep{delfosse2024adaptive}---the ratio of two polynomials with trainable coefficients---have shown promise in improving plasticity in standard RL and continual learning settings. Moreover, they should better mitigate unit saturation, avoiding issues like neuron death in ReLU or saturation in Tanh. However, their impact on stability remains unclear.

At the same time, recent studies~\citep{nauman2024overestimationoverfittingplasticityactorcritic} have identified continuous control environments that inherently promote instability in RL, particularly under high-update-to-data (UTD) training regimes, where policies are updated multiple times per data sample. These environments are prone to overestimation errors and gradient explosions, raising a crucial question: Does the increased flexibility of trainable rational activations help stabilize RL training in such settings, or does it instead amplify existing instabilities?
In this paper, we show that while rational activation functions enhance expressivity, their unconstrained flexibility can indeed exacerbate instability in high-UTD RL due to excessive activation scaling, leading to severe overestimation errors. To mitigate this, we propose a constrained variant of rational activations that introduces structural safeguards, preventing uncontrolled growth while retaining sufficient adaptability. Across both continuous and discrete RL tasks, we observe that continuous control environments—such as those in MetaWorld and DMC—exhibit particularly pronounced instabilities with unconstrained rational activations, especially under high update-to-data regimes. These settings reveal critical failure modes, including activation explosions and overestimation, which are less prevalent in discrete domains like Atari. This contrast helps isolate the conditions under which expressivity–stability trade-offs become most relevant.

Our experiments demonstrate that these constrained activations significantly improve stability in continuous control tasks, where standard rational activations struggled. However, we also observe a trade-off: while constrained rational activations improve stability, they sacrifice some of the improved plasticity of unconstrained rational activations. This highlights a fundamental balance between stability and adaptability in RL architectures.

To quantify these effects, Figure \ref{fig:dmc-mw-combined-subset} compares our constrained rational activations against standard approaches in MetaWorld and DeepMind Control Suite (DMC) environments. Our results show that introducing structural constraints enhances overall performance and reduces instability, though periodic resets remain necessary in high-UTD settings.

We make the following key contributions:
\begin{itemize}
    \item Identifying failure cases of current methods: We show that standard rational activations can lead to overestimation and other optimization pathologies in high-UTD RL, a failure mode not previously reported.
   \item     Proposing a constrained activation variant: We introduce modifications that improve stability without significantly limiting model flexibility. Our design is guided by principles similar to those used in initializing and regularizing standard activation functions.
   \item     Empirical validation: We demonstrate that our constrained activations significantly enhance RL training stability in continuous control environments while highlighting a trade-off with plasticity in continual learning.
   \item     Broader analysis: We explore the role of coefficient initialization, weight decay, and gradient covariance properties in shaping learning dynamics.
\end{itemize}

\textbf{Key Takeaway:}

Our study highlights that trainable activations must be carefully designed to balance expressivity and stability in RL. While increased flexibility can enhance adaptability, it can also introduce instability—necessitating constraints that ensure robust training dynamics. These insights provide a new perspective on activation function design and its broader implications for learning in non-stationary environments.

\section{Background and Related Work}

Activation functions are crucial in deep reinforcement learning (RL), shaping neural networks' representational capacity and optimization dynamics. While ReLU and Tanh are the most common choices~\citep{mnih2013playing,mnih2015human,hessel2018rainbow},  they contribute to various problems of learning dynamics~\citep{nauman2024overestimationoverfittingplasticityactorcritic}. In particular, ReLU is prone to dormant neurons phenomenom~\citep{sokar2023dormant}, plasticity loss~\citep{pmlr-v202-lyle23b,lyle2023understanding}, and overestimation on out-of-distribution data~\citep{DBLP:conf/icml/BallSKL23}. Tanh is often used to constrain the output, but it tends to saturate at its extremes, causing vanishing gradients that slow down learning in deep networks~\citep{pascanu2012difficulty}. Recently, Fourier features~\citep{tancik2020fourier} and learned Fourier features~\citep{li2021functional,DBLP:conf/iclr/YangAA22} emerged as efficient solutions for fitting complex value functions with high-frequency components. While Fourier features can be effective at fitting complex functions, even in a continual supervised learning setting \cite{lewandowski25_plast_learn_deep_fourier_featur}, they often suffer from overfitting and generalize worse on states with observation noise~\citep{mavor-parker2024frequency}.

Activation functions based on the Padé approximation, i.e., rational functions~\citep{DBLP:conf/iclr/0001SK20}, have been introduced to address some of these issues by allowing the activation function shape to be learned through backpropagation during training. \citet{delfosse2024adaptive} demonstrated that rational activation functions (RAF) are effective in RL, specifically for DQN~\citep{mnih2013playing,mnih2015human} and Rainbow~\citep{hessel2018rainbow} algorithms on Atari dataset~\citep{bellemare2012arcade}. In particular, their findings demonstrate that RAF mitigates overestimation issues and improve plasticity in a setting with image-based observation and discrete actions, but primarily in the low update-to-data (UTD) regime~\citep{d'oro2023sampleefficient,DBLP:conf/icml/SchwarzerOCBAC23}. Our work, in contrast, focuses primarily on
%a setting with continuous state-based observations, but in 
the high UTD regime, for which plasticity issues are more severe~\citep{DBLP:conf/icml/NikishinSDBC22}.
%with the off-policy algorithm Soft Actor-Critic~\citep[SAC,][]{haarnoja2018soft}.

\section{Instability of Rational Activation Functions}

%\subsection{Trainable Rational Activation Functions}
\label{sec:original_rationals}
In our study, we explore trainable rational activation functions, where the polynomial coeffecients are optimized alongside the rest of the network. Specifically, we follow the general form previously considered in the context of reinforcement learning~\cite{delfosse2024adaptive}, which is given by:
\[
f(x) = \frac{a_nx^n + a_{n-1}x^{n-1} + ... + a_1x + a_0}{|b_mx^m| + |b_{m-1}x^{m-1}| + ... + |b_1x| + 1}
\]  
where the coefficients, $\{a_i\}_{i=0}^n$ and $\{b_j\}_{j=1}^m$, are trainable. The absolute values in the denominator ensure numerical stability while maintaining flexibility.

Prior work~\cite{molina2020padeactivationunitsendtoend} suggests that selecting an appropriate degree for the numerator ($n$) and denominator ($m$) is crucial for achieving a balance between expressivity and stability. Based on our initial investigation (Appendix~\ref{appendix:degree-analysis}), we adopt the ($n=3$, $m=2$) configuration, which provides strong plasticity while avoiding numerical instability.

\subsection{Empirical Evidence of Instability}
\label{sec:limitations}

We aim to investigate whether rational activation functions, such as Padé approximants~\cite {molina2020padeactivationunitsendtoend,delfosse2024adaptive}, are universally effective across different applications and could serve as a replacement for static functions like ReLU. Our initial investigation (Appendix~\ref{appendix:degree-analysis}) reveals that rational activation functions may suffer from performance degradation over time, primarily due to the instability introduced by large activation coefficients, which can lead to high-magnitude losses.

Given these observations, we extended our investigation to reinforcement learning tasks, where input distributions are highly dynamic. 
Our results below demonstrate that rational activations lead to training instability in some DeepMind Control Suite~\citep{tassa2018deepmind} and MetaWorld~\citep{Yu2018MetaWorld} tasks.

\begin{figure}[h!]
    \centering  
    \includegraphics[width=0.4\linewidth,height=0.25\linewidth]{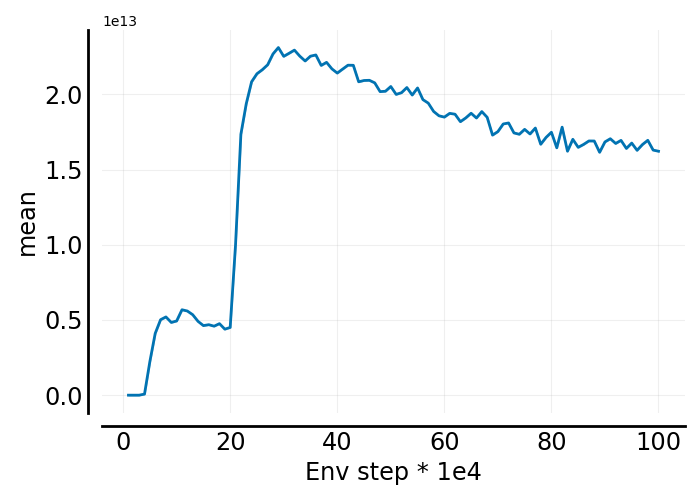}
    \includegraphics[width=0.4\linewidth,height=0.25\linewidth]{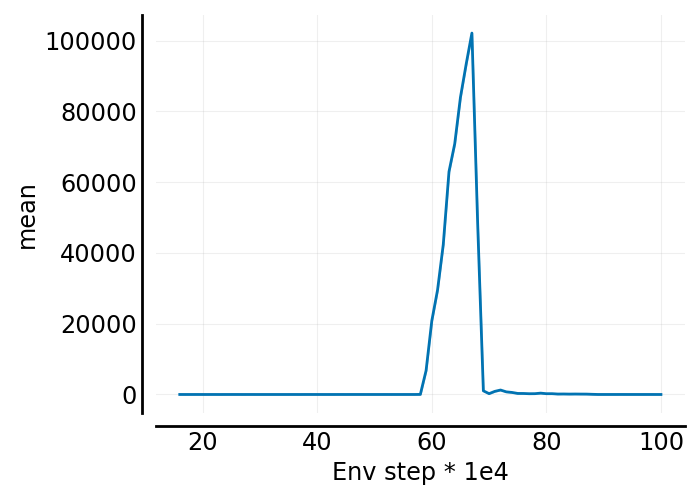}
    \includegraphics[width=0.4\linewidth]{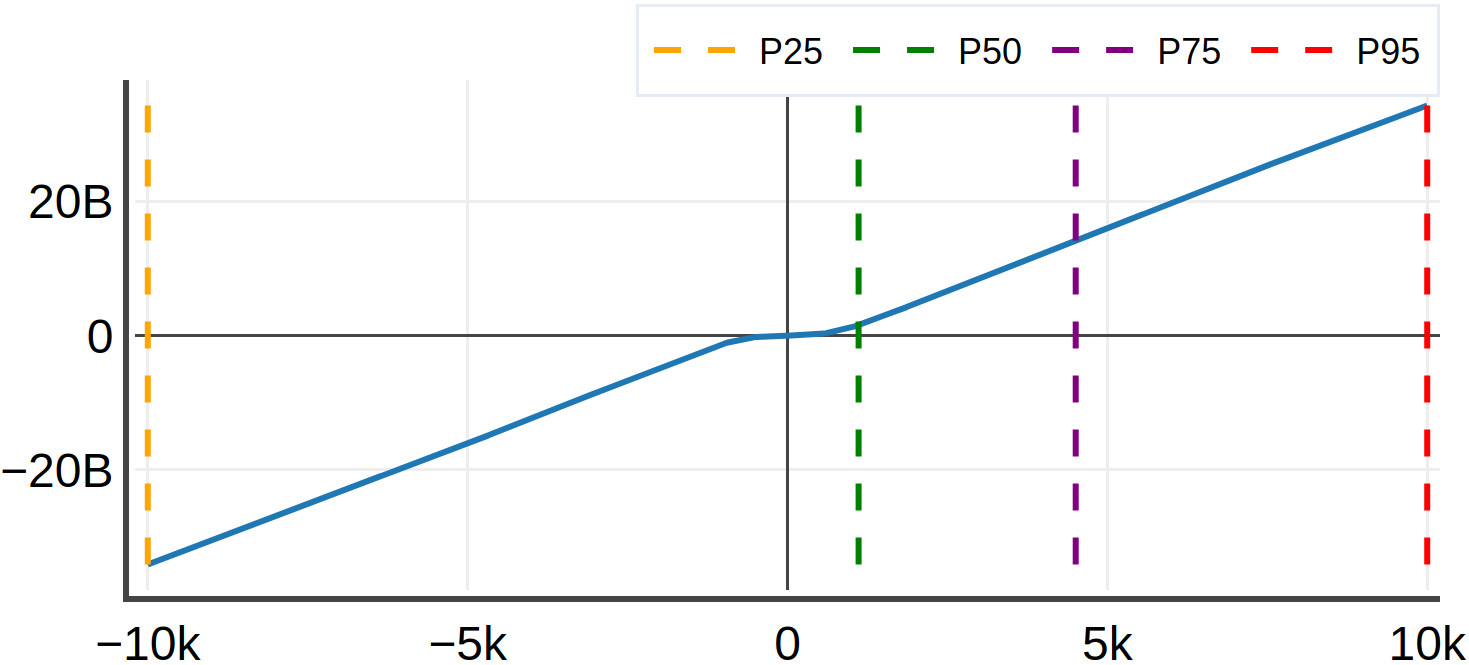}  
    \includegraphics[width=0.4\linewidth]{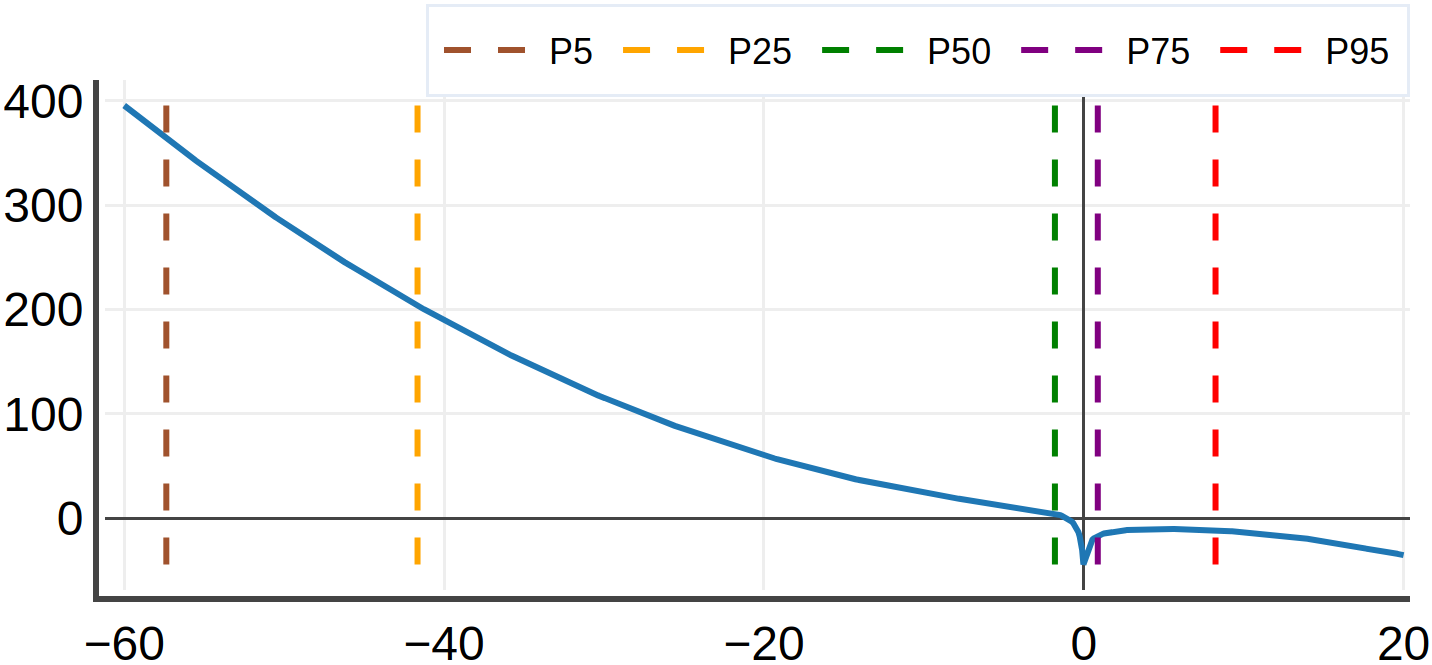}  
    \caption{Critic overestimation and activation distributions from the second hidden layer of critic networks trained with Soft Actor-Critic (SAC) under UTD of 10 after approx. 650K environment steps. 
    % The left plots corresponds to training in the Sweep environment, while the right plot shows activations from the Dog-Trot environment. 
    The bottom plot shows the rational activation function after training in
    Sweep (left):
    \( \frac{3.86x^3 + 123x^2 - 375x - 1114}{|0x^2| + |0x| + 1} \)
    , and Dog-Trot (right):
    \( \frac{-0.9x^3 + 6.1x^2 - 88x - 99}{|0x^2| + |-9.3x| + 1} \).
    Vertical dashed lines indicate the P5, P25, P50, P75, and P95 percentiles of the preactivation distribution in the second hidden layer, showing the range of inputs received by the activation function. Unconstrained rational activations produce large outputs, leading to severe overestimation.}
    \label{fig:dmc-mw-rationals}
\end{figure}

\paragraph{Experimental setup}
Reinforcement learning tasks, particularly those in DeepMind Control Suite (DMC) and MetaWorld, are known to involve unstable input distributions (\cite{nauman2024overestimationoverfittingplasticityactorcritic}).
To systematically analyze the behavior of rational activation functions in reinforcement learning, we conducted experiments using the Soft Actor-Critic (SAC)~\citep{haarnoja2018soft} algorithm, along with its various implementation variants.

The SR-SAC~\citep{doro2022sampleefficient} algorithm served as our baseline. In all experiments, both the critic network and the MLP used in the actor consisted of two hidden layers, where standard activations were replaced with trainable rational functions of degrees ($n=3$, $m=2$). Rational functions were initialized to imitate Leaky ReLU in the range $[-5,5]$. This setup allowed us to isolate the impact of activation function choice on training stability. In the most of our reinforcement learning experiments, we use a high UTD of 10, but we also examine how the different UTD regimes affect the behavior of rational activation functions. All the experiments were repeated on 5 random seeds.

\paragraph{Results}

\begin{wrapfigure}[14]{r}{0.4\linewidth}%[<liczba linii>]{<pozycja>}{<szerokość>}
    \centering
    \includegraphics[width=\linewidth]{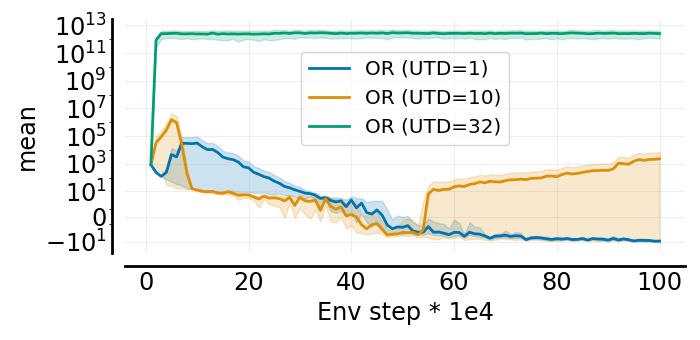}
    \caption{Mean overestimation on DMC benchmark for Original Rationals for different UTD settings. The higher UTD the less stable OR become leading to overestimation.}
    \label{fig:dmc-utd-overestimation}
\end{wrapfigure}

We found that original rational activation functions are highly sensitive to changes in the replay ratio. As UTD increases, instabilities emerge, leading to overestimation (Figure \ref{fig:dmc-utd-overestimation}). Using moderate UTD of 10, 
under which we conducted the most of our experiments, the instability of rational activation functions was particularly noticeable in MetaWorld environments. In many of them, such as Sweep, training the agent did not yield any improvements.

Even though we used a relatively stable configuration with low-degree rational functions, we observed that parameters in the numerator often grew significantly larger than those in the denominator. This imbalance caused the rational activation function to output extremely large values. These high-magnitude outputs, in turn, led to overestimation in the critic’s value predictions and training instability.

 This phenomenon resembles known instability mechanisms in deep networks, such as gradient explosion, where excessively large activations cause gradients to grow uncontrollably during backpropagation. Although our activations do not directly cause vanishing or exploding gradients in the classic sense, their unbounded growth under dynamic input distributions (as in reinforcement learning) indirectly induces similar failure modes—e.g., poor value estimation, divergent learning curves, and unstable policies. In particular, as shown in Appendix~\ref{appendix:rl-training}, Figure~\ref{fig:mw_gradnorms}, we observe that in MetaWorld environments, original rational functions (with or without resets) consistently exhibit extremely high gradient norms compared to constrained rational variants and ReLU. This suggests that excessive activation scaling leads to high-magnitude gradients, making this instability closely related to classic gradient explosion.

We refer to this behavior as activation explosion, characterized by excessive activation outputs due to coefficient imbalance. In MetaWorld environments, such as Sweep, this instability was severe and consistent. In contrast, DMC environments exhibited these issues more rarely—likely due to lower reward scales—but they still emerged under high update-to-data (UTD) regimes or when LayerNorm was applied, as seen in the Dog-Trot example in Figure~\ref{fig:dmc-mw-rationals}.

Interestingly, we did not observe similar activation explosions in discrete-action domains such as Atari, even under high replay ratios (e.g., UTD 32). This suggests that the instability phenomena we describe are specific to continuous control settings. A detailed analysis of rational activations on Atari 100k is provided in Appendix~\ref{appendix:atari-results}.

\paragraph{Takeaway}  

Our findings indicate that while rational activation functions can be powerful, they are not universally stable. The risk of instability is particularly high when:
\begin{itemize}
    \item The polynomial in the numerator is several degrees higher than in the denominator.
    \item The task involves highly dynamic inputs, such as reinforcement learning environments. 
    \item The network is trained under high UTD regime.
\end{itemize} 
Such circumstances can lead to a situation where the numerator coefficients become disproportionately large compared to the denominator (or the denominator coefficients converge to 0) and the denominator does not limit the numerator causing uncontrolled growth in function outputs. We define this phenomenon of activations producing excessively large outputs, leading to training failure, as \textit{explosions}. These results emphasize the need for additional regularization mechanisms when using rational activation functions in tasks with high input variability.

\section{Refining Rational Activation Functions with Structural Constraints}
%\subsection{Rational Activation Function with Internal Regularization}

As demonstrated in the previous section, certain configurations of rational activation functions lead to instability, particularly in reinforcement learning settings. To address these issues, we 
propose a more stable formulation of the rational activation function, and evaluate its effectiveness across different environments.

Specifically, we consider a rational activation function with structural constraints,
\begin{equation}
f(x) = \frac{a_nx^n + a_{n-1}x^{n-1} + \dots + a_1x}{|b_mx^m| + |b_{m-1}x^{m-1}| + \dots + |b_1x| + 1 + \left|\frac{x}{c}\right|^d}.
\end{equation}

This formulation introduces two key modifications:
\begin{itemize}
    \item \textbf{Internal Regularization via \(\left|\frac{x}{c}\right|^d\) in the denominator.} To prevent uncontrolled growth, we ensure that the highest power of \( x \) in the denominator exceeds that of the numerator (\( d > n \)). In our experiments, we set \( d = n+1 \), which forces the activation function to asymptotically decay to zero as \( |x| \to \infty \). This modification limits extreme outputs while preserving flexibility within a controlled input range.
    \item \textbf{Removal of the constant term \( a_0 \) from the numerator.} This ensures that \( f(0) = 0 \), aligning the function's behavior with common activation functions such as ReLU, Tanh, and Sigmoid. Empirically, we found that including \( a_0 \) often led to instability during training (see Appendix~\ref{appendix:stability_analysis}).
\end{itemize}

These modifications impose structural constraints that stabilize training while maintaining the adaptability of rational activations. By preventing excessive growth for large inputs and ensuring a consistent response near zero, our approach mitigates key weaknesses of unconstrained rational activations.

\subsection{Evaluation Setup: Addressing Training Challenges in RL and CL}

To comprehensively evaluate our constrained rational activation functions, we assess their performance in two distinct contexts:
\begin{wrapfigure}[20]{r}{0.5\linewidth}%[<liczba linii>]{<pozycja>}{<szerokość>}
    \vspace{-10pt}
    \centering
    \includegraphics[width=\linewidth]{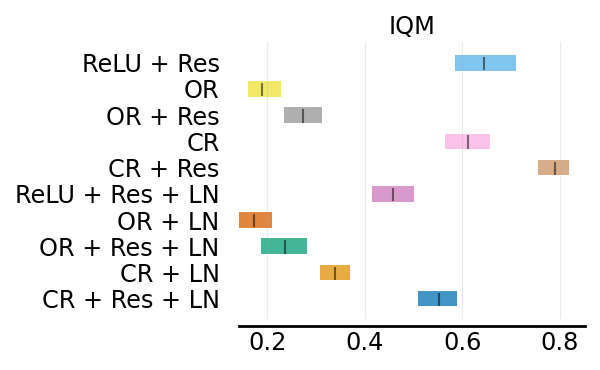}
    \caption{IQM performance after 1M steps, aggregated across 15 environments from MetaWorld (MW) and 15 from DeepMind Control Suite (DMC). DMC scores are rescaled by a factor of 1000 to match the range of MW scores. The plot compares various activation function configurations, including Original Rationals (OR), Constrained Rationals (CR), and ReLU, along with their combinations with Layer Normalization (LN) and Resets (Res). }
    \label{fig:dmc-mw-combined}
\end{wrapfigure}
\begin{itemize}
    \item Reinforcement Learning Training Dynamics – examining whether our modifications resolve the instability issues observed in prior experiments.
     \item  Continual Learning Adaptability – testing if our changes negatively impact plasticity over long training horizons.
\end{itemize}

For reinforcement learning experiments, we employ the Soft Actor-Critic (SAC) algorithm with different activations and evaluate their interaction with stabilization techniques such as periodic resets and Layer
Normalization~(\cite{ba2016layernormalization}). We examine rationals with degrees ($n=3$, $m=2$) and initialized as Leaky ReLU. We conducted our experiments across a diverse set of control environments, including tasks from DeepMind Control Suite~(\cite{tassa2018deepmindcontrolsuite}) and MetaWorld~(\cite{yu2021metaworldbenchmarkevaluationmultitask}). Specifically, we evaluated 15 distinct tasks from each framework, ensuring diversity in terms of the dimensionality of the observation and action spaces and the range of values for inputs and outputs.

For continual learning, we analyze the impact of our modifications on plasticity using the MNIST with non-stationary targets, i.e. where the labels are shuffled throughout training. Here, we study how different activation function initializations affect performance across 60 tasks. We specifically investigate whether the stability gains observed in RL come at the cost of plasticity loss in a continual learning setting.
By considering both settings, we aim to determine whether our proposed activation functions effectively address RL instabilities while preserving adaptability in CL scenarios.

\subsection{Results: Reinforcement Learning Training Dynamics}
\label{sec:dynamics}
\begin{wrapfigure}[16]{r}{0.4\linewidth}%[<liczba linii>]{<pozycja>}{<szerokość>}
    \vspace{-11pt}
    \centering
    \includegraphics[width=\linewidth]{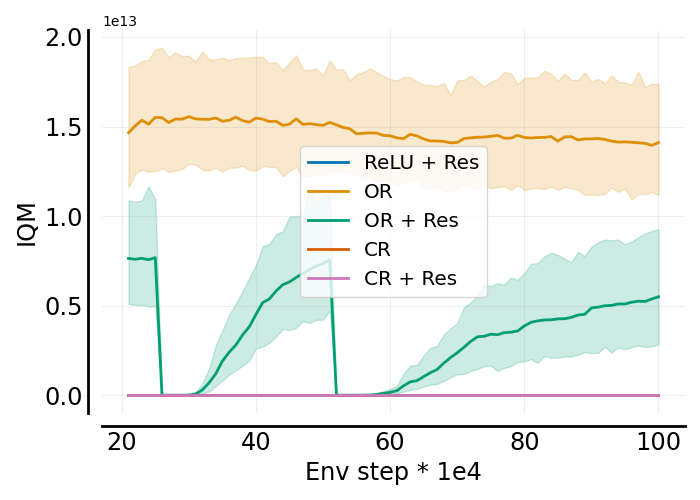}
    \caption{Overestimation on MW benchmark of different SAC setups. Original rational activation functions exhibit tendency to huge overestimation}
    \label{fig:mw-overestimation}
\end{wrapfigure}
Figure~\ref{fig:dmc-mw-combined} presents the combined results from MW and DMC. Since the performance scale differs between these frameworks, we normalized the reward values by dividing DMC rewards by 1000. Several key observations can be made:
\begin{itemize}
    \item Our constrained rational activation functions, when combined with periodic resets, achieve significantly better performance than networks using ReLU.
    \item Our constrained rational activation functions perform comparably to ReLU with resets.
    \item Original rational activation functions (even with resets) yield significantly worse results than other approaches.
\end{itemize}

Analyzing the performance of different activation functions across frameworks, we observe differences in the behavior of the original rational activation functions in DMC and MW. In DMC, agents using the original rational activation functions exhibit performance similar to that of agents using ReLU. On the other hand, we find that agents with original rational activation functions fail in MW environments. Deeper analysis shows that original rational activation functions tend to explode during training in multiple MW environments (Appendix \ref{appendix:rl-training}), leading to significant overestimation (Figure \ref{fig:mw-overestimation}).
Our modified rational activation functions eliminate the explosion problem present in the original functions. They perform well across both frameworks, achieving results comparable to ReLU with resets. Applying parameter resets in networks with rational activation functions generally improves agent performance. This effect is evident for both our modified version and the original functions in DMC. 
\begin{figure}[h!]
    \centering  
    \includegraphics[width=0.4\linewidth]{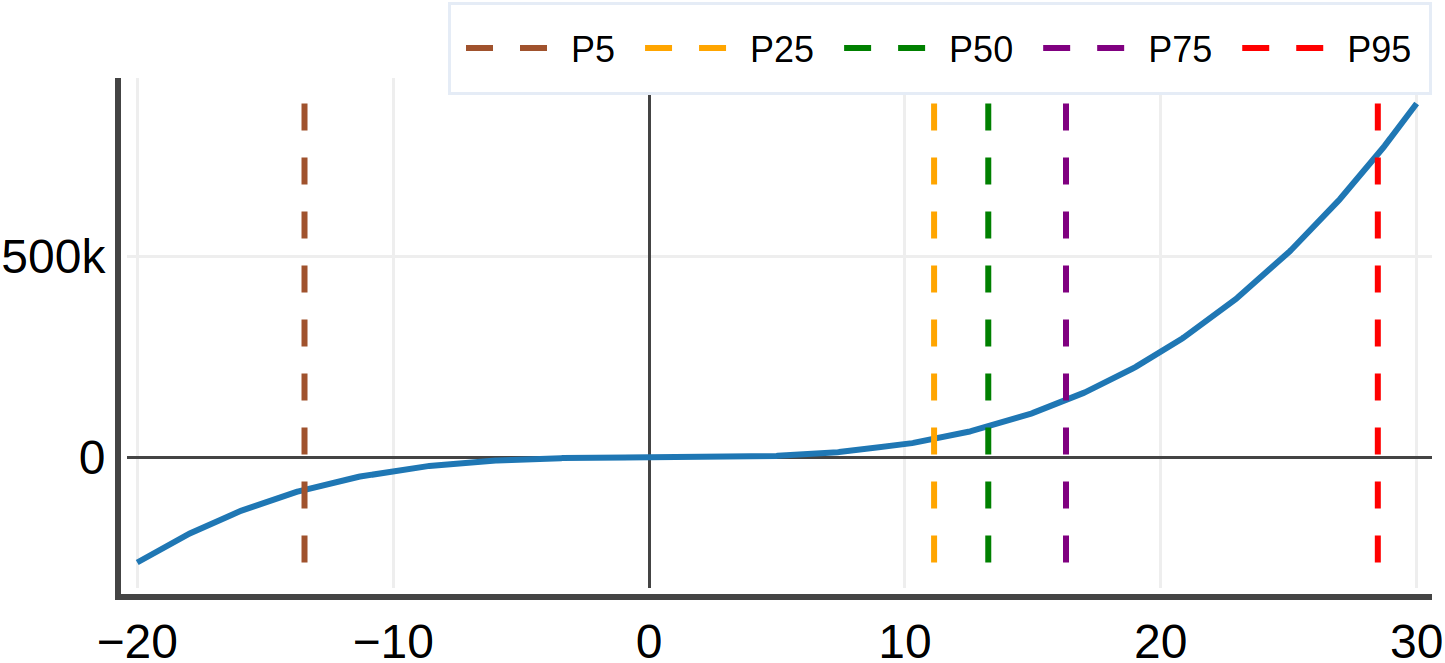} 
    \includegraphics[width=0.4\linewidth]{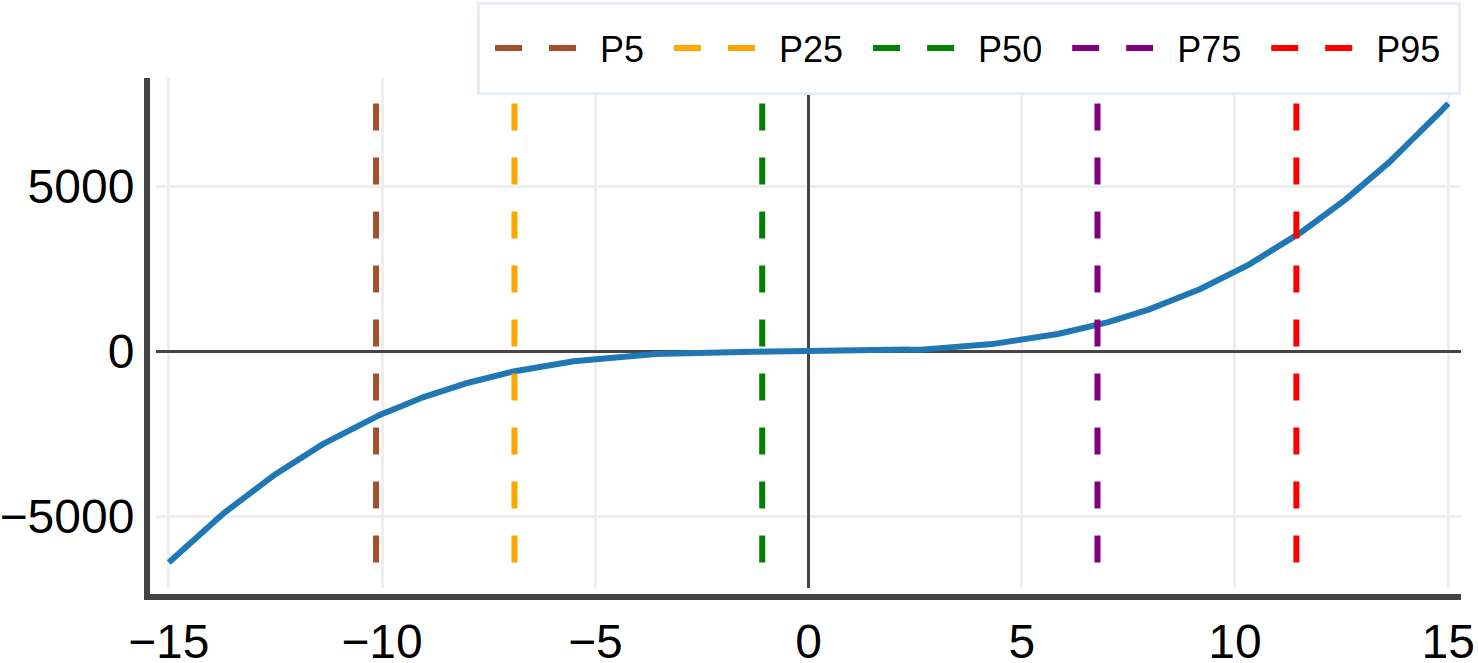}  
    \caption{Activation distributions from the second hidden layer of critic networks using LayerNormalization trained with Soft Actor-Critic (SAC) after approx. 500K environment steps in Sweep (left), with learned formula
    \( \frac{65x^3 + 2,4x^2 + 59x + 11,6}{|0x^2| + |0x| + 1} \),
    and Dog-Trot (right), with learned formula \( \frac{4,4x^3 + 5,4x^2 + 2,7x - 5}{|0x^2| + |0x| + 1} \).
    Even though LN keeps pre-activations close to 0 the output values are large, which leads to significant overestimation.
    }  
    \label{fig:dmc-mw-rationals-ln-explosion}  
\end{figure}
Additionally, rational activation functions make better use of individual samples and reach higher reward regions more quickly (Appendix~\ref{appendix:rl-performance-profiles}). This trend is evident both at the beginning of training and immediately after a network parameter reset. However, resets do not mitigate the explosion problem of the original rational functions in MW. There is also a conclusion that rational activation functions indeed mitigate the problem of network plasticity loss to some extent, though they do not eliminate it entirely thus still need resetting.

Intuitively, one might expect that Layer Normalization (LN) would help keep pre-activations distributions close to zero (addressing the issue illustrated in Figure \ref{fig:dmc-mw-rationals}) and thus stabilizing rational activations. 
\begin{wrapfigure}[10]{l}{0.4\linewidth}%[<liczba linii>]{<pozycja>}{<szerokość>}
    \vspace{-10pt}
    \centering
    \includegraphics[width=\linewidth]{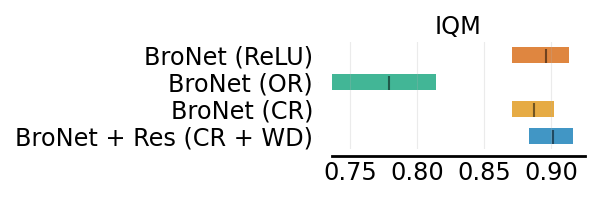}
    \caption{BroNet with different activation functions on DMC \& MW combined results.}
    \label{fig:dmc-mw-bro}
\end{wrapfigure}
However, we discovered that this is not the case. Rational activation functions still exhibit explosive behavior (Figure \ref{fig:dmc-mw-rationals-ln-explosion}). Adding LN introduces instabilities in the training of rational activation functions. In environments where explosions were previously absent, they began occurring after LN was applied (especially in MW). Original rational activation functions are particularly vulnerable to this problem. We find that our regularization reduces this effect on DMC envs where the results of our constrained rationals stay the same, contrary to OR which performs significantly worse when combined with resets.

In our experiments, we also examined how rational activation functions perform in networks with higher capacity, which currently achieve the best results in RL (\cite{nauman2024biggerregularizedoptimisticscaling}) (\cite{lee2024simbasimplicitybiasscaling}). We repeated the previous experiments but replaced the critic network with a BroNet architecture (\cite{nauman2024biggerregularizedoptimisticscaling}). 
Figure \ref{fig:dmc-mw-bro} presents the aggregated results for DMC and MW. Note that increasing network capacity increases expressivity, which has the effect of limiting the benefit of increased expressivity from trainable activation functions. However, even in this case, the original rational activation functions caused training instabilities in MW environments. 
Our modified rational activation functions perform slightly worse in the initial training phases, but do not introduce regressions in the final performance. Adding weight decay to the activation coefficients helps keep the performance high.

% \paragraph{Continual Learning Adaptability }
\subsection{Results: Continual Learning Adaptability}
 \begin{wrapfigure}[16]{r}
 {0.45\linewidth}
 \vspace{-12mm}
    \centering
    \includegraphics[width=0.89\linewidth]{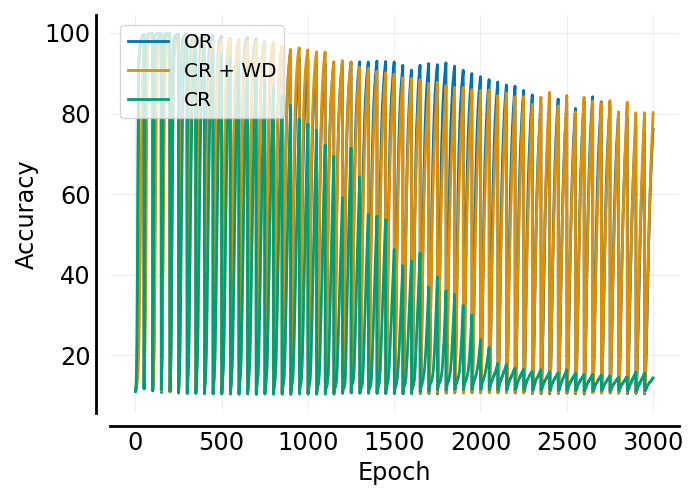}
    \caption{MNIST with labels reshuffling over 60 tasks: Original rational functions retain plasticity longer than modified version. WD helps our Constrained Rationals preserve adaptability.}
    \label{fig:plasticity_ratio_12}
\end{wrapfigure}
To evaluate the adaptability of our modified rational activation functions in a continual learning scenario, we utilize the experimental setup described in Section~\ref{sec:original_rationals}. Specifically, we train a model on the MNIST dataset with randomly assigned labels, periodically reshuffling the label assignments. This setup, commonly used to measure plasticity loss in neural networks, assesses the ability of the network to adapt to evolving target distributions~\cite{lewandowski2024curvature, lewandowski2024learning, lyle2022understanding, kumar2023maintaining, pmlrv202sokar23a}.

Analyzing Figure~\ref{fig:plasticity_ratio_12}, we find that while the constrained rational activation functions (CR) retain plasticity to some extent, they degrade faster than the original rational functions (OR). This suggests that our modifications while stabilizing RL training, introduce some rigidity that limits long-term adaptability. Once more, weight decay is applied to activation coefficients, which keeps a good performance for all of the tasks. Further inspection of this continual learning setup and weight decay is discussed in Section~\ref{sec:ratio_space_analysis}.

To complement this analysis, we also evaluate adaptability on a more structured continual learning benchmark, Split/CIFAR-100, in Appendix~\ref{appendix:split-cifar}. There, our constrained rational functions outperform ReLU and original rationals on later tasks, highlighting their ability to retain useful representations even as task distributions evolve.

\subsection{Trade-off Between Training Dynamics Stability and Plasticity}  
\begin{wrapfigure}[14]{r}{0.4\linewidth}
\vspace{-1.3mm}
\centering
    \includegraphics[width=\linewidth]{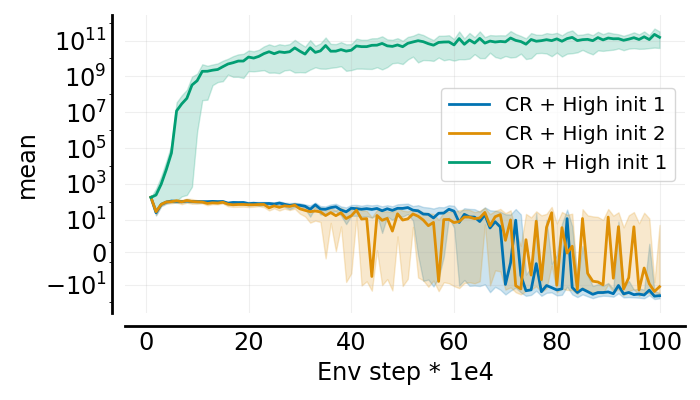}
    \caption{MetaWorld: Overestimation comparison between Original Rationals (OR) and Constrained Rationals (CR) with large initialization scales. OR remains less stable and can lead to severe overestimation, while CR shows more controlled behavior, mitigating instability.}
    \label{fig:mw-inits-overestimation-iqm}
\end{wrapfigure}
Bringing together insights from reinforcement learning and continual learning experiments, we observe a fundamental trade-off in our modifications. On the one hand, our improved rational activation functions resolve instability issues in RL scenarios with proprioceptive state inputs, leading to superior performance. On the other hand, these same modifications hinder the ability of the network to maintain plasticity in continual learning setups. This indicates that while our approach successfully mitigates the challenges observed in RL training, it comes at the cost of plasticity, raising further questions about balancing stability and plasticity in neural networks.

 \label{sec:initialization}
\begin{figure}[ht]
    \centering
    \includegraphics[width=0.45\linewidth]{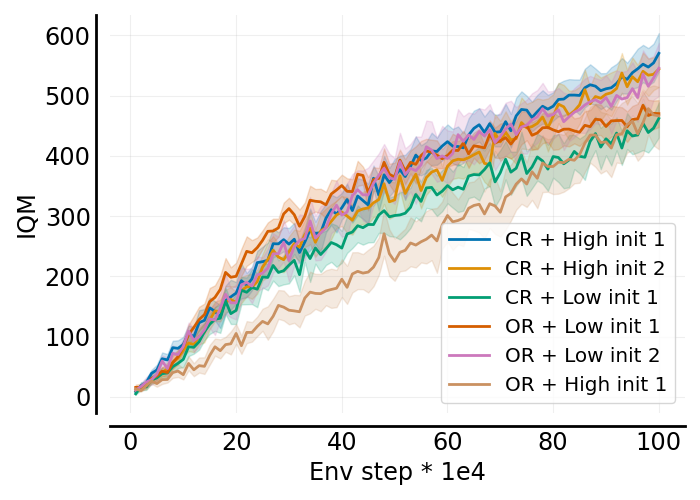}
    \includegraphics[width=0.45\linewidth]{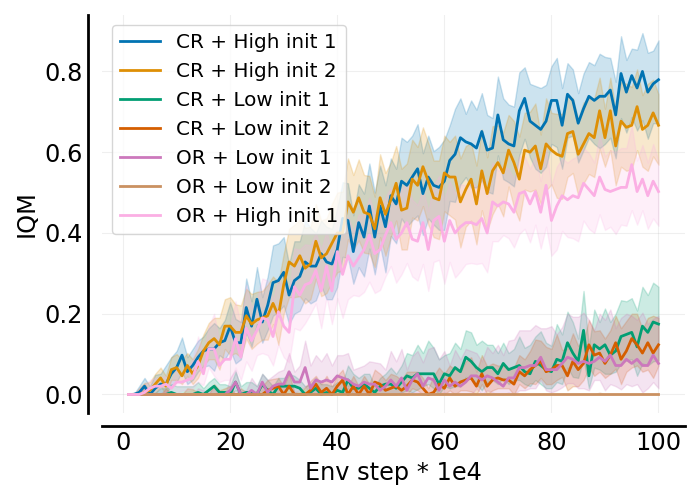}
    \caption{Left: DMC: Impact of coefficient initialization on RL performance. No clear trend is observed.
    Right: MetaWorld: Only activations initialized with large coefficients successfully solve tasks.}
    \label{fig:mw-dmc-inits-iqm}
\end{figure}
 \section{Analyzing the Space of Rational Activations}
 \label{sec:ratio_space_analysis}

Our modifications to rational activation functions have stabilized reinforcement learning (RL) training but reduced robustness to discontinuous task changes in continual learning (CL). To understand this trade-off, we examine how different factors influence training dynamics, focusing on coefficient initialization, weight decay, and Neural Tangent Kernel (NTK) properties.

\subsection{The Effect of Coefficient Initialization}

Unlike standard activations (e.g., ReLU), trainable activation functions—particularly rational functions—adapt their shape during training. This flexibility makes their initialization crucial for learning dynamics, affecting both stability and plasticity. Here, we examine:

\begin{itemize}
    \item How initial coefficient magnitudes impact network adaptability.
\item Differences between our modified rational activations and the original formulation.
\item Insights from Neural Tangent Kernel (NTK)~\citep{jacot2018neural} analysis.
\end{itemize}

\vspace{-2mm}
\textbf{Reinforcement Learning: DMC and MetaWorld}
In reinforcement learning, the impact of coefficient initialization is less straightforward. In the DeepMind Control (DMC) suite, no clear trend emerges regarding whether small or large coefficients provide better performance (Figure~\ref{fig:mw-dmc-inits-iqm}). However, in MetaWorld, a strong pattern is observed: only functions initialized with high coefficients successfully solve tasks, whereas those initialized with low values suffer from instability and overestimation (Figure~\ref{fig:mw-dmc-inits-iqm}).

A key observation is that while high initial coefficients improve stability in RL, they do not entirely eliminate overestimation, as seen in Figure~\ref{fig:mw-inits-overestimation-iqm}. This suggests that while larger coefficients increase expressivity and robustness to gradient explosion, they do not address fundamental instability issues inherent in rational functions without further regularization.

\begin{figure}[h!]
    \centering
    \includegraphics[width=0.4\linewidth]{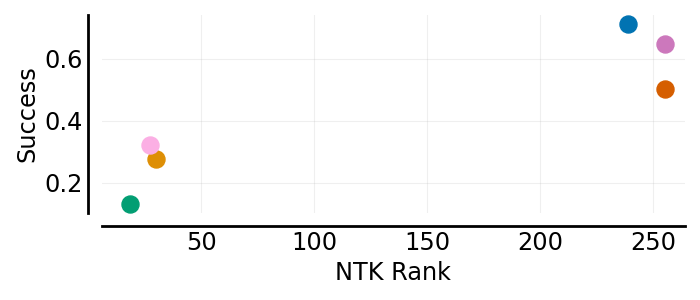}
    \includegraphics[width=0.4\linewidth]{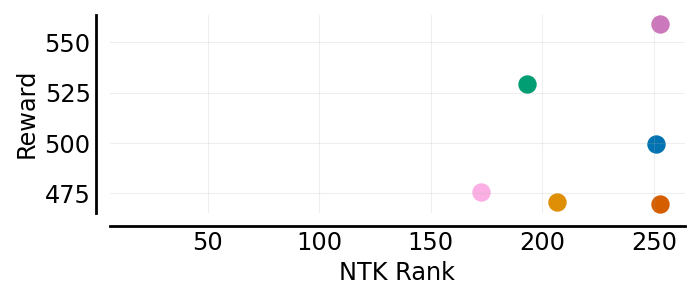}
    \raisebox{0.4\height}{\includegraphics[width=0.15\linewidth]{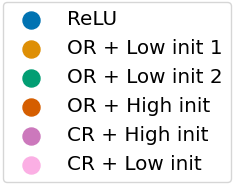}}
    \caption{ NTK analysis for different coefficient initializations in RL. Higher coefficient values increase NTK rank in MetaWorld (left), while DMC (right) maintains a high rank regardless of initialization.}
    \label{fig:ntk-rl-placeholder}
\end{figure}
\textbf{NTK Analysis}  
%To gain further insight, we examine the NTK spectrum in RL settings (Figure~\ref{fig:ntk-rl-placeholder}). Our analysis indicates that higher coefficient values lead to an NTK with greater rank and eigenvalue spread, suggesting more substantial feature separation. This may explain the improved stability observed in high-coefficient MetaWorld experiments. Conversely, small coefficients lead to a more degenerate NTK, making the model susceptible to instability in high-update-to-data RL settings.
To better understand the dynamics introduced by rational activation functions, we examine the Neural Tangent Kernel (NTK) in reinforcement learning settings (Figure~\ref{fig:ntk-rl-placeholder}). Intuitively, the NTK reflects how parameter updates influence model predictions during training. A higher-rank NTK with a broader eigenvalue spectrum suggests greater feature separation and expressivity, which can contribute to more stable learning — particularly important in environments with high input variability. Our analysis reveals that larger rational coefficients tend to induce this richer NTK structure, which may explain the improved stability observed in high-coefficient MetaWorld experiments. In contrast, smaller coefficients produce a more degenerate NTK, leaving the model more susceptible to instability under high update-to-data (UTD) regimes.

\begin{figure}[h!]
    \centering

    \includegraphics[width=0.4\linewidth]{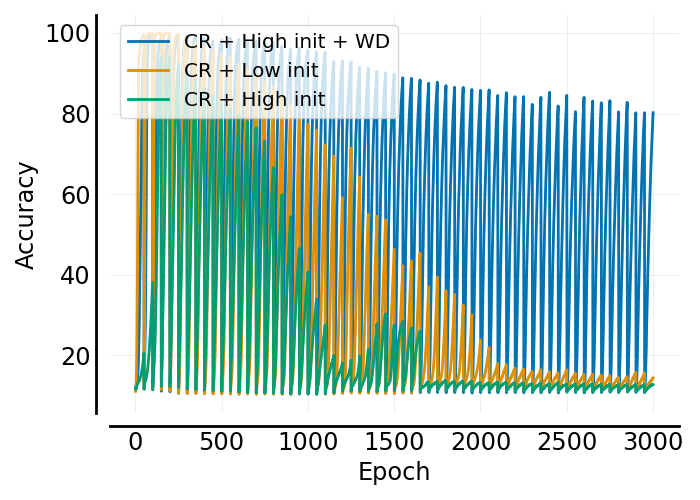}
    \includegraphics[width=0.4\linewidth]{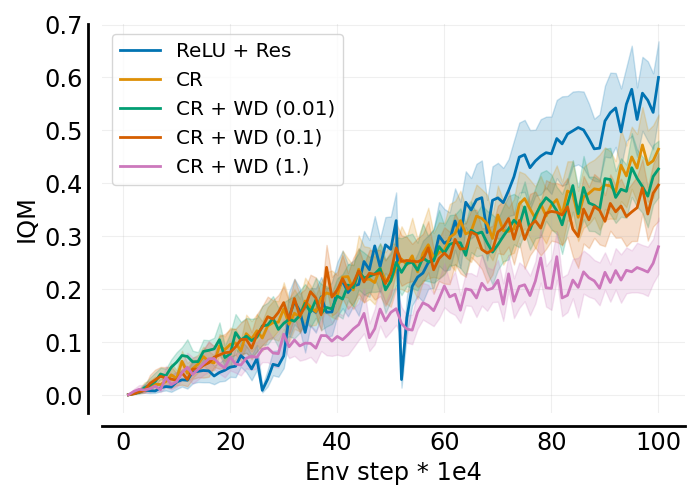}
    \includegraphics[width=0.4\linewidth]{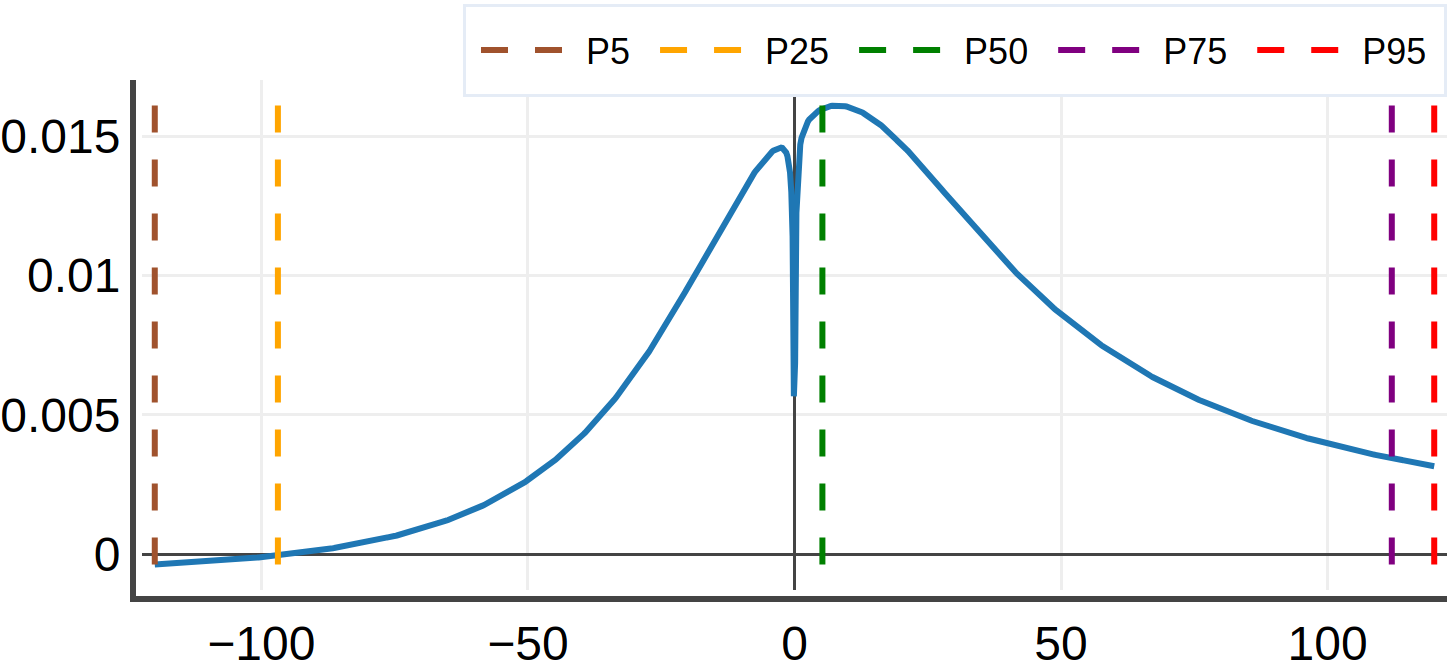}
    \includegraphics[width=0.4\linewidth]{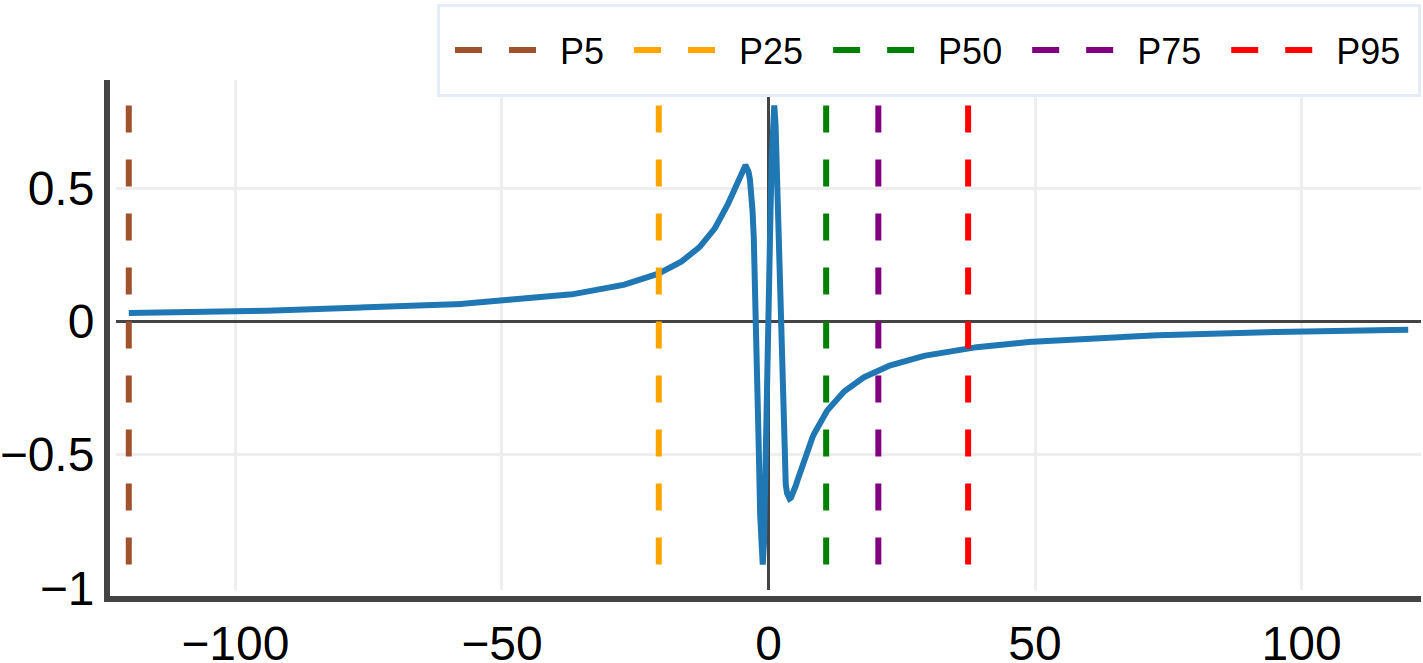}
    \caption{Impact of weight decay applied exclusively to rational activation coefficients. In top row. Left: Performance in a continual learning setup using Reshuffled-labels MNIST. Right: Performance in a reinforcement learning setup. This RL performance plot is aggregated across both DeepMind Control Suite (DMC) and MetaWorld (MW) environments. Bottom. Activations plot without WD (left) and with WD (right). Without applying WD, denominator coefficients grew significantly, causing the function to output low values. WD reduces this effect, keeping coefficients low and stabilizing the output. Functions formulas:
    Left:
    \( \frac{0.015x^3 + 1.38x^2 + 0.016x}{|87x^2| + |7.49x| + 1 +  |\frac{x}{2}|^4} \)
    , Right:
    \( \frac{-0.24x^3 - 0.05x^2 + 1.14x}{|0x^2| + |0x| + 1 + |\frac{x}{2}|^4} \)
    }  
    \label{fig:wd-performance}
\end{figure}
Interestingly, in DMC, NTK analysis reveals that the rank of the NTK remains relatively high across all initializations. This explains the lack of a clear performance trend in Figure~\ref{fig:mw-dmc-inits-iqm}; since the function space remains expressive regardless of initialization, the network's ability to learn is not significantly affected by the magnitude of the coefficients.
We provide the NTK analysis for different time points during training for selected environments, along with visualizations, in Appendix~\ref{appendix:cl-experiments}.

\textit{Takeaway:} While higher coefficient values improve feature separation and stability, NTK expressivity in DMC suggests that initialization effects are environment-dependent.

\textbf{Continual Learning}
We investigate the effect of coefficient initialization on plasticity in a continual learning setting using the Permuted MNIST benchmark. Our results indicate that networks initialized with smaller coefficient values exhibit significantly higher plasticity throughout training. This effect is particularly evident in more extended continual learning scenarios, where our modified rational activation functions struggle to retain adaptability compared to the original formulation.

To gain further insight, we analyze the evolution of the Neural Tangent Kernel (NTK) spectrum under different initialization schemes. We find that for Orignal Rationals lower coefficient values correspond to a reduced NTK rank, suggesting a more constrained function space that better preserves plasticity. In contrast, our Constrained Rationals have high NTK rank no matter the initialization. This suggests that our regularization enhances expressivity but accelerates feature collapse over multiple tasks. A detailed breakdown of these findings, along with visualizations, is provided in Appendix~\ref{appendix:rl-ntk-heatmaps}.

\textit{Takeaway:} Smaller coefficient values improve plasticity in continual learning by constraining the function space, while higher values enhance expressivity at the cost of faster feature collapse.

\vspace{-1mm}
\subsection{Weight decay}
\vspace{-1mm}
Our proposed Constrained Rationals improve stability in reinforcement learning but exhibit reduced plasticity in continual learning (CL) tasks. As shown in Figure~\ref{fig:wd-performance}, networks with rational activations in CL suffer from a gradual loss of adaptability, which appears to be linked to changes in the learned activation coefficients. Specifically, we observe that the denominator coefficients in Constrained Rationals tend to increase significantly while the numerator coefficients decrease. This results in a compression of the activation range, meaning that the function outputs increasingly smaller values for a given input range.

A key issue arises from coefficient drift and pre-activation shifts. As pre-activations move further from zero, our Internal Regularization term \( \left|\frac{x}{c}\right|^d \) causes activations to decay toward zero, reducing neuron responsiveness and plasticity.  

Without weight decay, denominator coefficients expand, pushing activations toward zero (Figure~\ref{fig:wd-performance}, bottom-left). Vertical lines mark pre-activation percentiles, showing many fall outside \([-100, 100]\), where activations vanish, leading to neuron inactivity. Applying weight decay (Figure~\ref{fig:wd-performance}, bottom-right) mitigates this drift, keeping activations within a nonlinear range and preserving adaptability.  

Weight decay, commonly used to prevent parameter drift, serves a similar role for rational activations -- when applied specifically to activation coefficients: constraining coefficient growth and preventing activation collapse. Figure~\ref{fig:wd-performance} (left) shows that this improves plasticity in CL, but in RL (Figure~\ref{fig:wd-performance}, right), strong regularization harms performance, suggesting excessive constraint limits learning efficiency in dynamic environments.

Interestingly, this trade-off between constraint and expressivity appears to be architecture-dependent. In higher-capacity networks such as BroNet~\citep{nauman2024biggerregularizedoptimisticscaling}, we observe that applying weight decay to activation coefficients does not impair performance—in fact, it slightly improves IQM metrics across MetaWorld and DMC environments (Figure~\ref{fig:dmc-mw-bro}). This suggests that in sufficiently expressive networks, modest regularization may retain benefits for long-term training stability without limiting adaptability, highlighting the importance of tuning regularization strategies to the model's capacity and task demands.

\textit{Takeaway:} Weight decay applied to activation coefficients prevents activation collapse and improves plasticity in CL. However, in RL, excessive regularization hampers performance, indicating a trade-off between stability and adaptability.

\vspace{-3mm}
\section{Conclusion}
\vspace{-2mm}
In this work, we investigated the effectiveness of trainable rational activation functions in reinforcement learning (RL) under high-update-to-data (UTD) settings. While previous studies suggested that rational activations improve plasticity in standard RL and continual learning (CL) benchmarks, our results reveal that the original unconstrained version can lead to instability in continuous control tasks. Specifically, we identified a previously unreported failure mode of the original unconstrained rational functions in that their instability leads to large activations that tend to exacerbate overestimation issues.

To address these challenges, we introduced a modified version of rational activations with structural constraints designed to balance expressivity and stability. Our empirical evaluation across MetaWorld and DeepMind Control Suite (DMC) tasks demonstrated that these constrained rational activations outperform both standard ReLU-based architectures and their unconstrained rational counterparts. However, we also observed that these modifications come at the cost of reduced adaptability in continual learning setups, highlighting a fundamental trade-off between stability in RL and long-term plasticity in CL.

These findings suggest that while trainable activation functions offer promising advantages, their design requires careful consideration depending on the learning paradigm. Future work should explore dynamic mechanisms for adjusting activation function expressivity throughout training, potentially enabling a more adaptive balance between stability and plasticity. Additionally, further theoretical analysis is needed to understand the impact of constrained rational activations on optimization landscapes and representation learning in deep RL.

%\subsubsection*{Author Contributions}

\subsubsection*{Acknowledgments}
This manuscript was partially supported by the National Science Centre, Poland, under a grant 2023/51/D/ST6/01609, and by Warsaw University of Technology within the Excellence Initiative: Research University (IDUB) programme.
 We gratefully acknowledge Polish high-performance computing infrastructure PLGrid (HPC Centers: ACK Cyfronet AGH) for providing computer facilities and support within computational grant no. PLG/2025/018111.
Moreover, we would like to thank Atish Agarwala, Michał Nauman, and Wojciech Masarczyk for helpful comments on the manuscript.

\bibliography{collas2025_conference}
\bibliographystyle{collas2025_conference}
\appendix
\section{Preliminary Analysis of Rational Activation Stability}
\label{appendix:degree-analysis}

\paragraph{Continual learning}
To investigate the stability of rational activation functions, we first conducted a controlled experiment on a continually changing MNIST classification task. A multilayer perceptron (MLP) was trained for 500 epochs, with label permutations applied every 50 epochs to simulate a non-stationary learning scenario. The objective was to assess how different numerator and denominator degrees influence plasticity and stability. 

\begin{figure}
\centering
\includegraphics[width=0.45\linewidth]{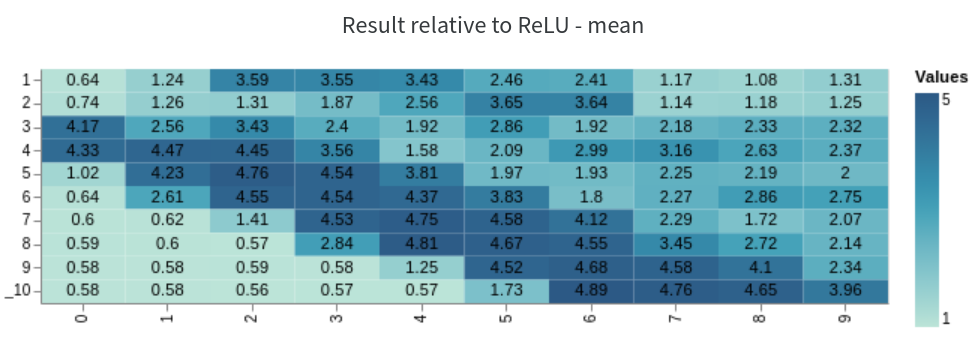}
\includegraphics[width=0.45\linewidth]{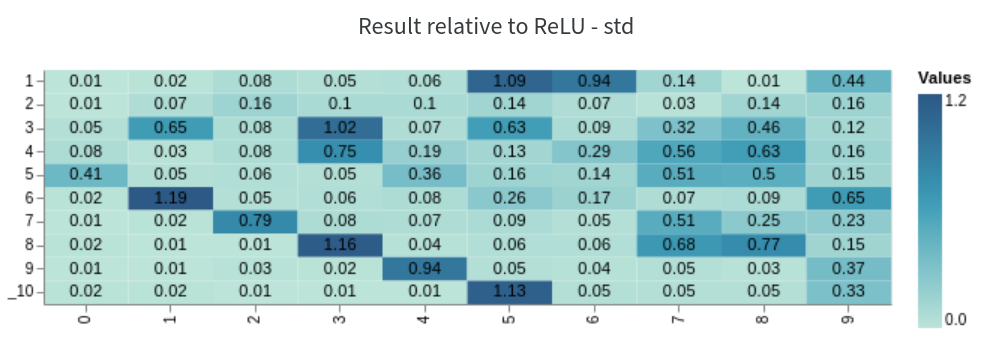}
\caption{Performance of trainable rational activation functions with varying numerator and denominator degrees on MNIST. Left: Mean AUC. Right: Standard deviation of AUC.}
\label{fig:heatmap-ratio}
\end{figure}

Figure~\ref{fig:heatmap-ratio} summarizes the results. The heatmap on the left shows the mean area under the curve (AUC) of training accuracy (normalized by ReLU performance), while the right panel depicts standard deviations, capturing variance in performance across different configurations. 

Our findings indicate that activation functions where the numerator degree slightly exceeds the denominator degree perform best. A higher-degree numerator enables greater expressiveness, while an excessively complex denominator often introduces instability. Among tested configurations, the coordinates (4,2) (i.e. ($n=3$, $m=2$) setting emerged as the most balanced, maintaining strong plasticity while avoiding numerical instability. 

\begin{figure}  
    \centering  
    \includegraphics[width=0.5\linewidth]{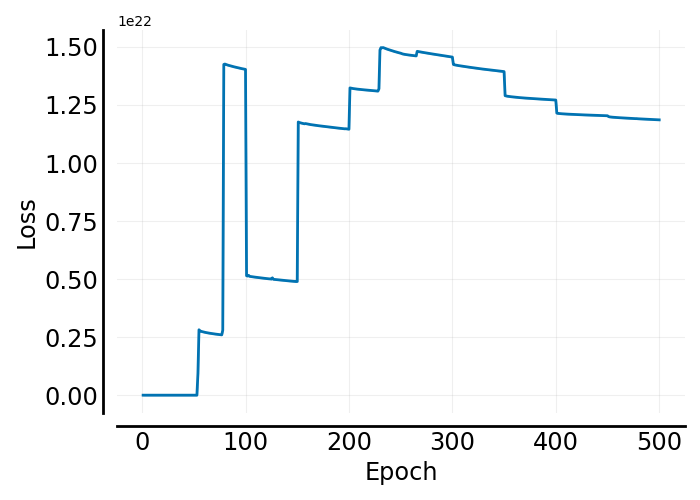}  
    \caption{Training loss for a (7,1) rational activation on MNIST, showing severe instability.}  
    \label{fig:pmnist-7-1-loss}  
\end{figure}

However, certain configurations led to severe instability. For instance, Figure~\ref{fig:pmnist-7-1-loss} illustrates the training loss for a (7,1) (i.e. ($n=6$, $m=1$)) rational activation, where an overpowered numerator combined with a weak denominator resulted in loss explosions. A similar issue was observed with purely polynomial activations (e.g., (5,0)), where the absence of a stabilizing denominator led to uncontrolled growth.  

\paragraph{Reinforcement learning}
These findings suggest that the degree balance in rational activations is critical for stability. Given that reinforcement learning (RL) environments feature highly dynamic input distributions, we hypothesized that similar instability may arise in such settings. This motivated our extension of the study to RL tasks, where we examine whether rational activations exacerbate instability in high-update-to-data (UTD) regimes. As demonstrated in Section~\ref{sec:limitations}, our results confirm that in some DeepMind Control Suite~\citep{tassa2018deepmind} and MetaWorld~\citep{Yu2018MetaWorld} tasks, unconstrained rational activations introduce severe instability, limiting their applicability in RL.

\begin{figure}
    \centering
    \includegraphics[width=0.45\linewidth]{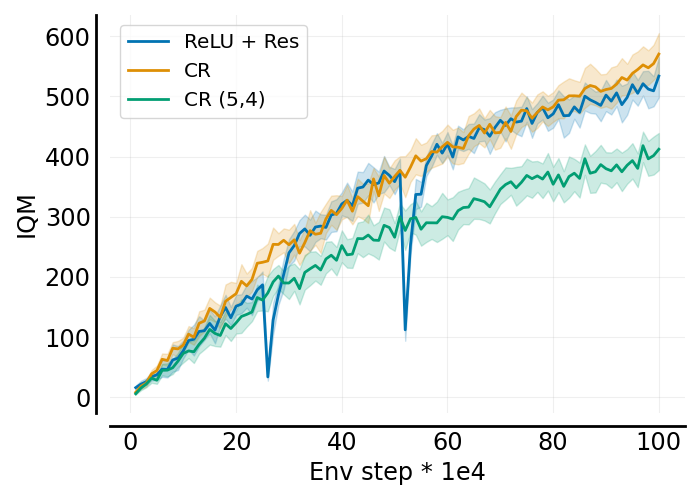}
    \includegraphics[width=0.45\linewidth]{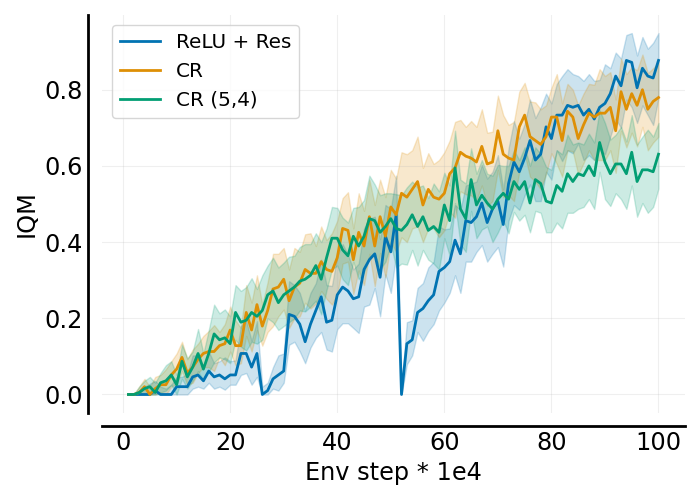}
    \caption{IQM performance of our Constrained Rationals with different polynomial degrees across 1M steps aggreegated over 15 tasks. Rewards for DeepMind Control Suite (left) and Success rate for MetaWorld (right).}
    \label{fig:dmc-mw-ratio-5-4}
\end{figure}

We also tested rational activation functions of degree (5,4) following the original paper \cite{delfosse2024adaptive}. The results are presented in Figure~\ref{fig:dmc-mw-ratio-5-4}. It is evident that increasing the polynomial degrees did not have a positive impact on the agent's performance.

\section{Reinforcement learning experiments}

\subsection{Hyperparameters}
In this section, we present the network hyperparameters used in all experiments. Table \ref{tab:sac-params} provides an overview of the parameters for the Soft Actor-Critic algorithm, while Table \ref{tab:ratio-params} contains the initial values of the rational activation functions coefficients. The coefficients Original Rationals Low init and Constrained Rationals High init were computed using the SciPy library to approximate the Leaky ReLU function over the range [-5, 5]. The remaining parameters were manually designed using function visualization tools. If multiple initializations of the same type were used in the text (e.g., Low init 1 and Low init 2), the second one was obtained by modifying the parameters from Table \ref{tab:ratio-params} to produce a different function shape while maintaining the same magnitude of coefficients.
\begin{table}[h]
    \centering
     \caption{Hyperparameters for SAC algorithm used in reinforcement learning experiments.}
    \begin{tabular}{ll}
        \toprule
        \textbf{Parameter} & \textbf{Value} \\
        \midrule
        Batch size ($B$) & 256 \\
        Replay ratio & 10 \\
        Critic hidden depth & 2 \\
        Critic hidden size & 256 \\
        Actor depth & 2 \\
        Actor size & 256 \\
        Actor learning rate & 3e-4 \\
        Critic learning rate & 3e-4 \\
        Temperature learning rate & 3e-4 \\
        Optimizer & ADAM \\
        Discount ($\gamma$) & 0.99 \\
        Initial temperature ($\alpha_0$) & 1.0 \\
        Exploratory steps & 10,000 \\
        Target entropy ($\mathcal{H}^*$) & $|\mathcal{A}|/2$ \\
        Polyak weight ($\tau$) & 0.005 \\
        \bottomrule
    \end{tabular}

    \label{tab:sac-params}
\end{table}

\begin{table}[h]
    \centering
    \caption{Example of a rational function parameter table}
    \begin{tabular}{lcccccc}
        \toprule
        \textbf{Rational type} & $a_3$ & $a_2$ & $a_1$ & $a_0$ & $b_1$ & $b_0$ \\
        \midrule
        Original Rationals Low init & 0.096 & 0.651 & 1.178 & 0.381 & 0.149 & 0.249 \\
        Original Rationals High init & 5.000 & 10.000 & 40.000 & 0.100 & 5.000 & 34.000 \\
        Constrained Rationals Low init & 0.500 & 0.100 & 2.000 & N/A & 0.050 & 0.500 \\
        Constrained Rationals High init & 3.966 & 19.190 & 15.780 & N/A & 0.000 & 34.800 \\
        \bottomrule
    \end{tabular}
    \label{tab:ratio-params}
\end{table}

\subsection{Constrained rationals stability analysis}
\label{appendix:stability_analysis}
\begin{figure}
    \centering
    \includegraphics[width=0.45\linewidth]{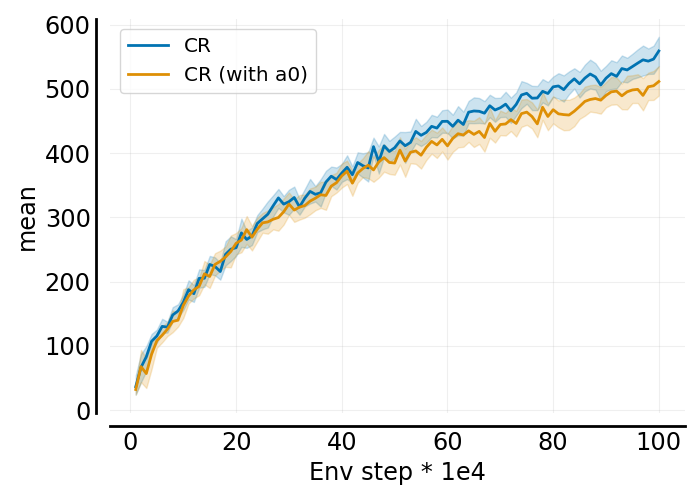}
    \includegraphics[width=0.45\linewidth]{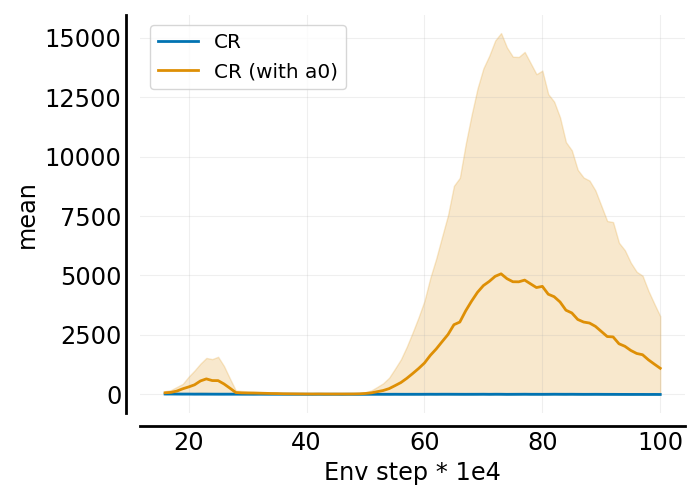}
    \includegraphics[width=0.45\linewidth]{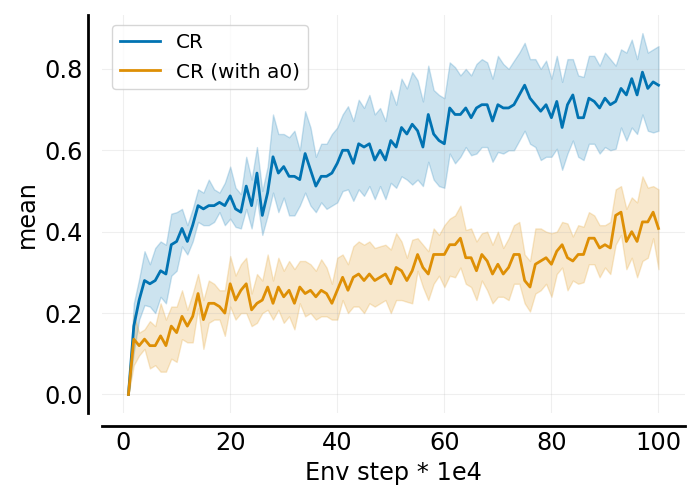}
    \includegraphics[width=0.45\linewidth]{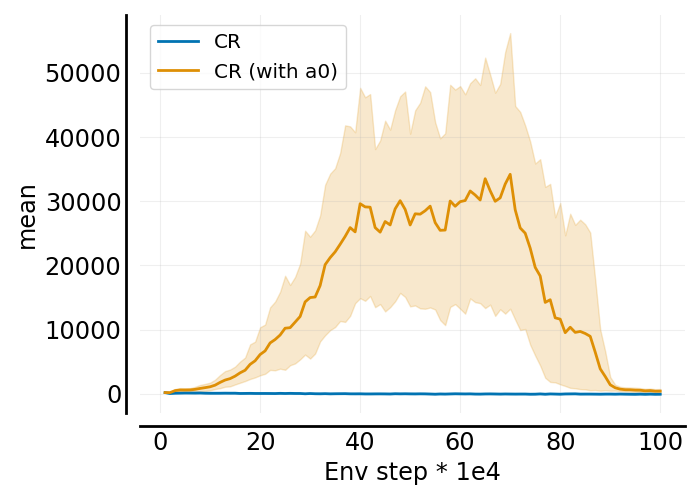}
    \caption{Comparison of Constrained Rationals (CR) with and without $a_0$ in the numerator. Left: scores, Right: overestimation. Top: DMC, Bottom: MW. It is clearly visible that constant term introduces overestimation during training and leads to lower results.}
    \label{fig:custom_ratio_with_vs_without_a0}
\end{figure}

Figure~\ref{fig:custom_ratio_with_vs_without_a0} illustrates the impact of including the trainable \( a_0 \) term in the numerator of Constrained Rationals. While its presence has a minor effect on performance in DeepMind Control Suite (DMC), it significantly degrades results in MetaWorld. Moreover, when analyzing overestimation, we observe that the inclusion of \( a_0 \) consistently leads to overestimation across both benchmarks, with the effect being notably more pronounced in MetaWorld compared to DMC.

\subsection{RL training dynamics}
\label{appendix:rl-training}
A deeper analysis of activation function behavior revealed the root cause of the explosions issue on MetaWorld. The higher order of magnitude of rewards obtained by the agent led to larger network losses. This, in turn, resulted in greater gradient magnitudes, which, when combined with the lack of regularization in original rational functions, caused rapid growth in activation function parameters. The increasing parameter values led to erroneous (excessively large) network outputs, further amplifying parameter growth. Ultimately, this resulted in complete training instability and a failure to improve rewards.

However, this issue does not occur with our proposed rational activation functions with regularization. Agents using these functions achieve only slightly lower rewards than those employing ReLU with resets. However, it is evident that rational activation functions perform worse in the later stages of training, where reward growth becomes stagnant. Adding resets to rational activation functions enables the network to reach high reward levels much faster, comparable to ReLU-based networks. Moreover, in this framework, the increase in rewards shortly after parameter resets is more pronounced, indicating more efficient sample utilization by networks using rational activation functions.

\begin{figure}
    \centering
    \includegraphics[width=0.5\linewidth]{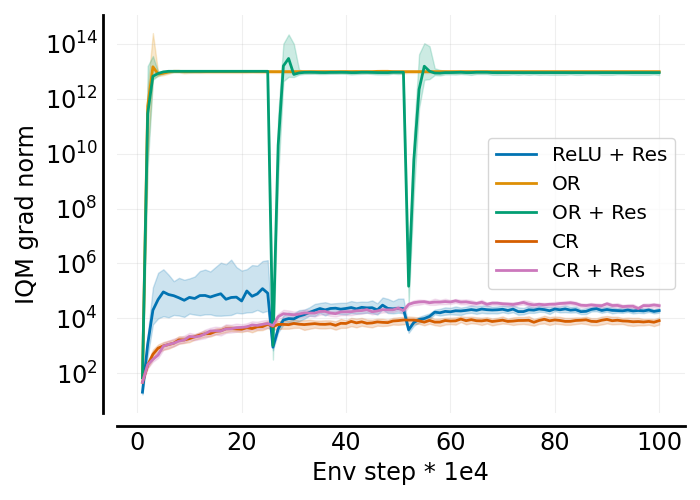}
    \caption{Gradient norms of activation coefficients during MetaWorld training with UTD 10. We compare Original Rationals (OR) with and without resets, Constrained Rationals (CR), and ReLU with resets.}
    \label{fig:mw_gradnorms}
\end{figure}
To better understand these training instabilities, we plot the gradient norms of activation coefficients during MetaWorld training in Figure~\ref{fig:mw_gradnorms}. The figure compares Original Rationals (OR), Constrained Rationals (CR), and ReLU (with resets). We observe that OR—with or without resets—exhibits clear signs of gradient explosion, as indicated by a rapid and unbounded increase in coefficient gradients. In contrast, CR shows consistently lower and more stable gradient norms, even without resets, highlighting its robustness. ReLU with resets maintains low gradient norms throughout, serving as a stable baseline. These results reinforce the need for structural constraints in trainable activations to ensure stable training dynamics in reinforcement learning.

\subsection{RL Performance Profiles}
\label{appendix:rl-performance-profiles}
In this section, we analyze agent performance over training timesteps to evaluate learning speed and stability across different activation function configurations.

\begin{figure}[h!]
    \centering
    \includegraphics[width=0.5\linewidth]{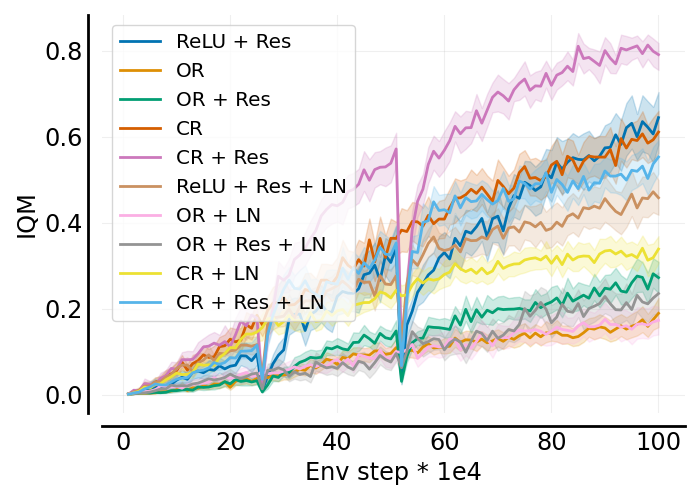}
    \caption{IQM performance across 1M steps, aggregated over 15 MetaWorld (MW) and 15 DeepMind Control Suite (DMC) environments. DMC scores are rescaled by a factor of 1000 to match MW score ranges. The plot compares activation function configurations: Original Rationals (OR), Constrained Rationals (CR), and ReLU, along with their combinations with Layer Normalization (LN) and Resets (Res).}
    \label{fig:dmc-mw-results}
\end{figure}

Figure~\ref{fig:dmc-mw-results} presents IQM performance across 1M steps, aggregated over 15 MetaWorld (MW) and 15 DeepMind Control Suite (DMC) environments. The results indicate that while Constrained Rationals (CR) and Original Rationals (OR) provide higher expressivity, they can suffer from instability, particularly in high-update-to-data (UTD) settings. The inclusion of Layer Normalization (LN) and Resets (Res) improves stability across both benchmarks.

\begin{figure}[h!]
    \centering
    \includegraphics[width=0.45\linewidth]{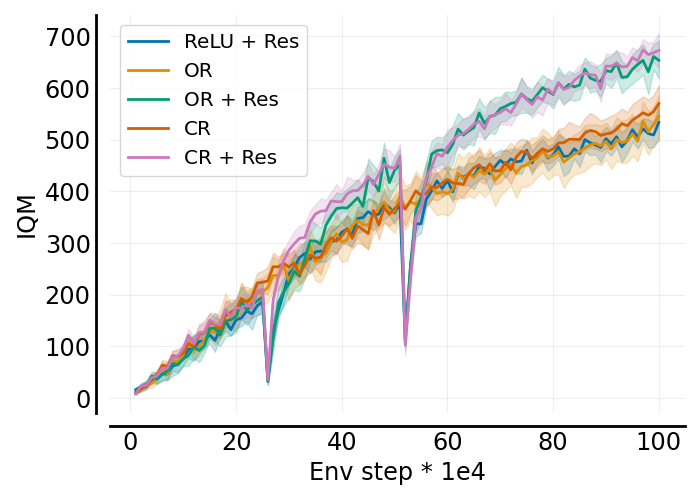}
    \includegraphics[width=0.45\linewidth]{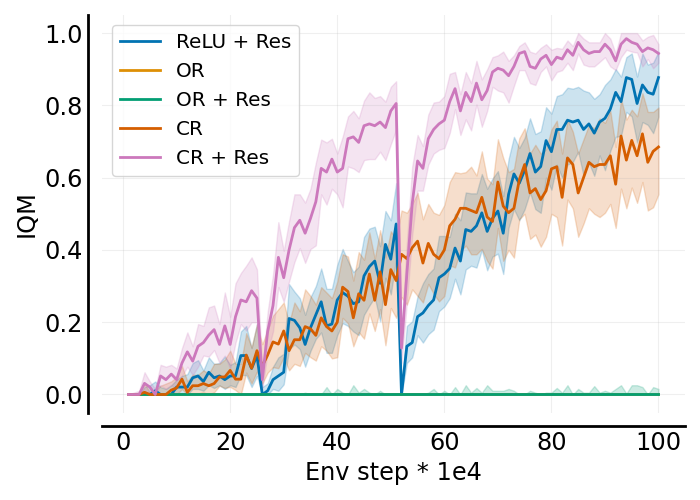}
    \caption{Performance across 1M steps. Left: DMC returns. Right: MW goals. Comparison of activation function configurations: Original Rationals (OR), Constrained Rationals (CR), and ReLU, with and without Resets (Res).}
    \label{fig:dmc-mw-results_rr10}
\end{figure}

To further dissect this behavior, Figure~\ref{fig:dmc-mw-results_rr10} separately examines DMC returns and MW goals, comparing various activation functions with and without Resets. We observe that rational activation functions exhibit increased variance in MW, suggesting that additional constraints may be necessary to improve consistency.

\begin{figure}[h!]
    \centering
    \includegraphics[width=0.45\linewidth]{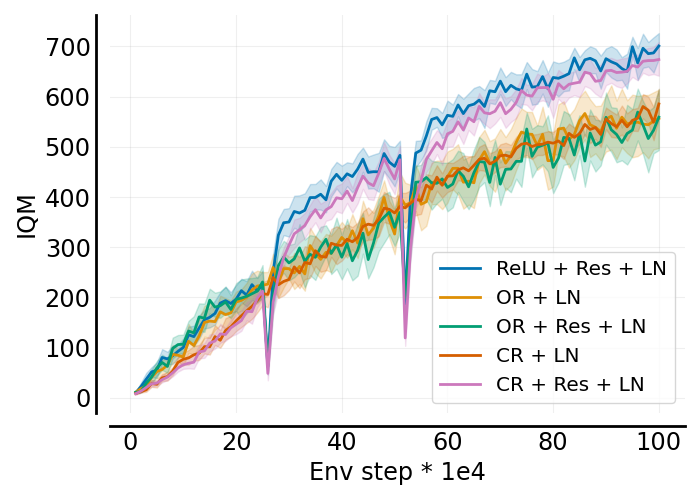}
    \includegraphics[width=0.45\linewidth]{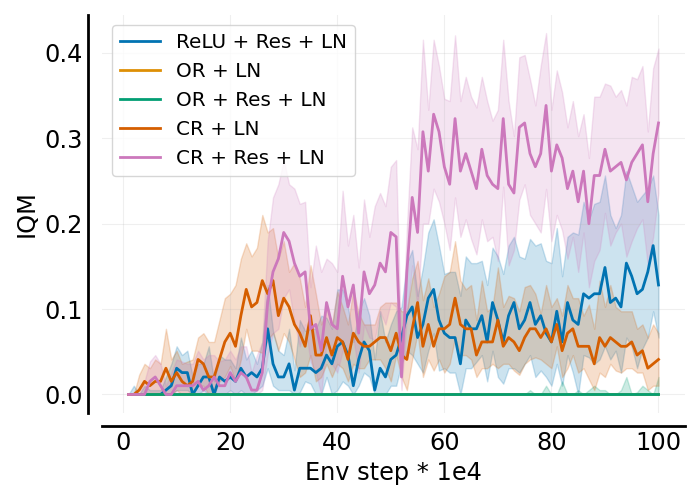}
    \caption{Performance across 1M steps. Left: DMC returns. Right: MW goals. Comparison of activation function configurations: Original Rationals (OR), Constrained Rationals (CR), and ReLU, with Layer Normalization (LN) and Resets (Res).}
    \label{fig:dmc-mw-results_ln}
\end{figure}

Figure~\ref{fig:dmc-mw-results_ln} extends this analysis by incorporating Layer Normalization (LN). Here, we find that the combination of LN and Resets helps mitigate instability, particularly in MW environments where overestimation errors are more pronounced.

\begin{figure}[h!]
    \centering
    \includegraphics[width=0.5\linewidth]{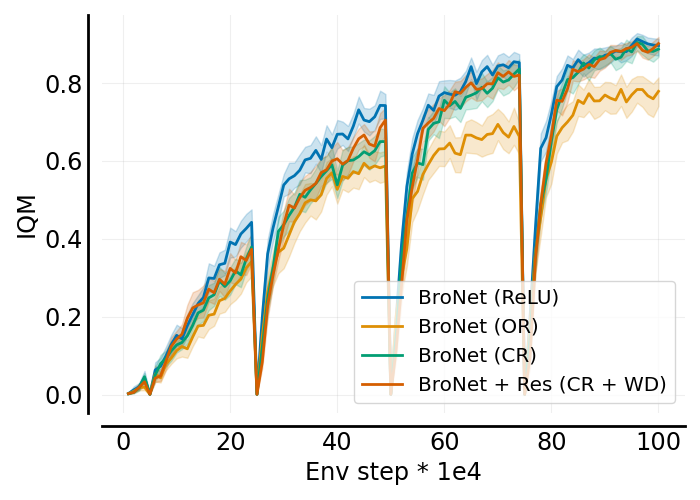}
    \caption{Comparison of BroNet architecture with different activation functions: Original Rationals (OR), Constrained Rationals (CR), and ReLU, aggregated over MW and DMC benchmarks.}
    \label{fig:dmc-mw-results_bro}
\end{figure}

Next, we evaluate performance within the BroNet architecture. Figure~\ref{fig:dmc-mw-results_bro} shows the aggregated results across MW and DMC benchmarks, comparing OR, CR, and ReLU activations. While BroNet provides improvements in certain tasks, rational activations still demonstrate sensitivity to coefficient drift.

\begin{figure}[h!]
    \centering
    \includegraphics[width=0.45\linewidth]{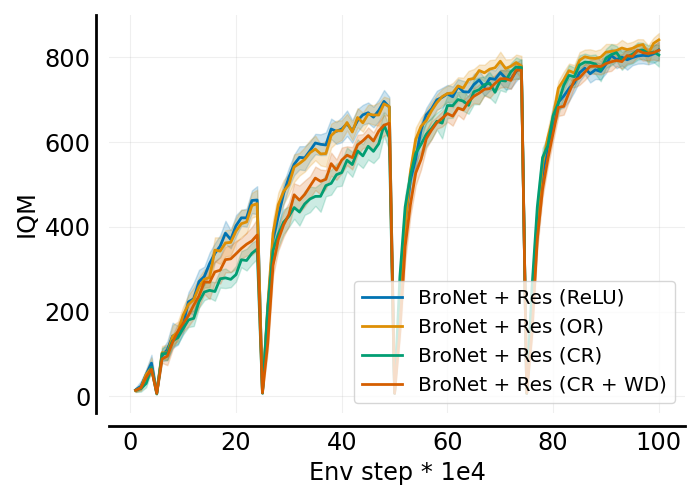}
    \includegraphics[width=0.45\linewidth]{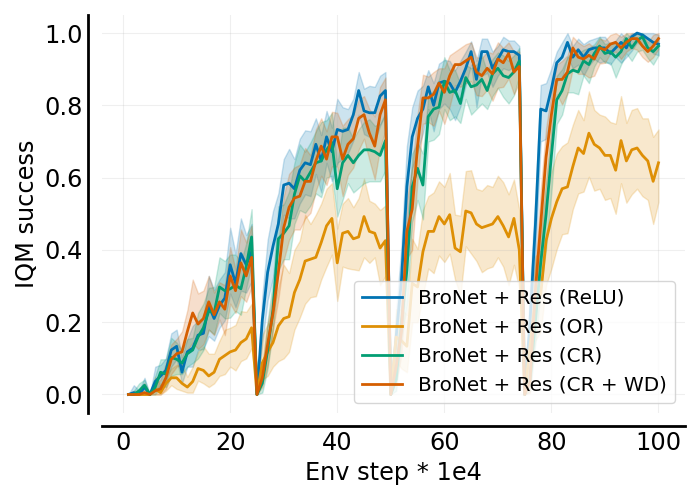}
    \caption{Performance of BroNet architecture across 1M steps. Left: DMC returns. Right: MW success rate. Breakdown of results from Figure~\ref{fig:dmc-mw-results_bro} into separate benchmarks.}
    \label{fig:dmc-mw-results_bro_split}
\end{figure}

A more detailed breakdown of these results is provided in Figure~\ref{fig:dmc-mw-results_bro_split}, which separates the BroNet results into DMC returns (left) and MW goals (right). We see that performance gains are more consistent in DMC, whereas MW remains challenging for rational activations due to persistent overestimation issues.

\subsection{Spectral Normalization in RL Training}
\label{appendix:sn-analysis}

To explore alternative regularization strategies beyond weight decay, we evaluated the impact of Spectral Normalization (SN) applied to the weight matrices of actor and critic networks across both MetaWorld (MW) and DeepMind Control Suite (DMC) benchmarks.

\begin{figure}[h!] 
\centering 
\includegraphics[width=0.45\linewidth]{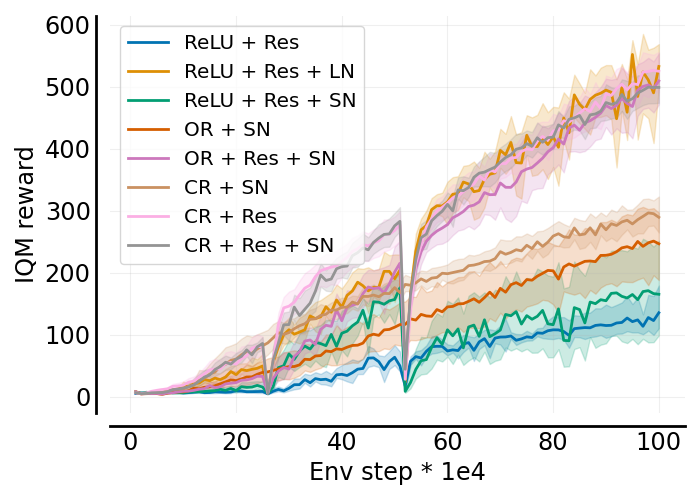} 
\includegraphics[width=0.45\linewidth]{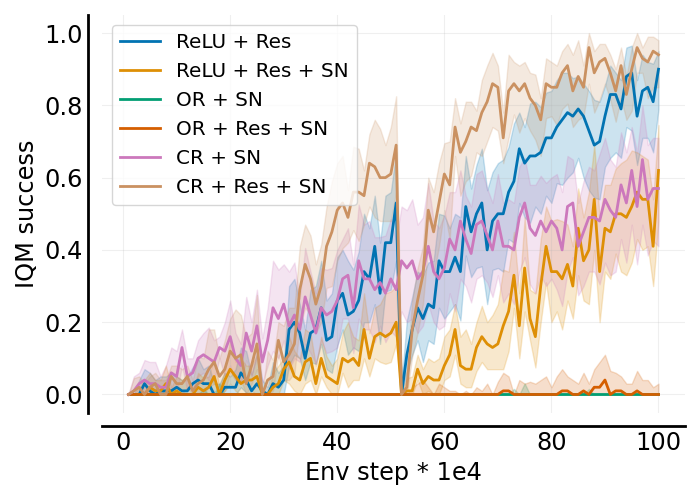} 
\caption{Impact of Spectral Normalization (SN) on RL performance across 1M steps under UTD=10. Left: DMC returns. Right: MW goals. We compare ReLU, Original Rationals (OR), and Constrained Rationals (CR), with and without Resets, combined with SN.} \label{fig:sn-rl-results}
\end{figure}

As shown in Figure~\ref{fig:sn-rl-results}, SN provides a noticeable improvement for Constrained Rationals (CR). Without resets, CR+SN outperforms both OR+SN and ReLU+SN across both DMC and MW benchmarks. When combined with resets, CR+SN achieves the strongest performance on MW and remains competitive with the best alternatives on DMC.

These findings suggest that SN complements the structural constraints of CR activations, particularly by enhancing training stability in environments prone to coefficient drift and activation explosion.

\subsection{Spectral Normalization on activations outputs}

To further investigate sources of instability in training networks with learnable activation functions, we explored the application of Spectral Normalization (SN) not only to weight matrices but also to the outputs of activation functions themselves. Intuitively, applying SN at this stage could reduce the detrimental effects of growing activation magnitudes, especially for expressive parametric functions such as rational functions.

This hypothesis was motivated by our observation that unconstrained rational functions (OR) often exhibit unbounded growth in output magnitude, which may exacerbate issues of exploding activations and lead to training collapse. Constrained Rationals (CR), by contrast, enforce structural limits that stabilize training, but we sought to determine whether SN could provide an alternative or complementary regularization mechanism.

\begin{figure}
    \centering
    \includegraphics[width=0.45\linewidth]{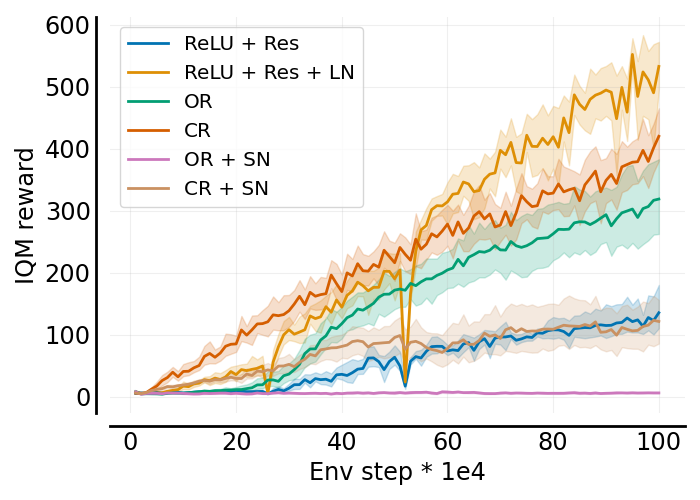}
    \includegraphics[width=0.45\linewidth]{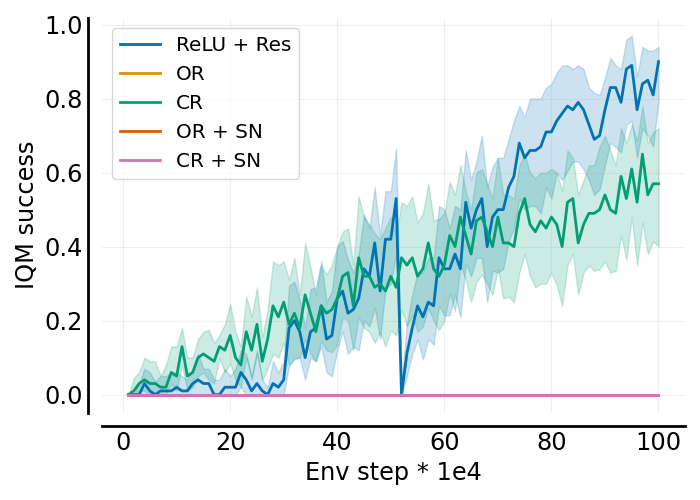}
    \caption{Impact of applying Spectral Normalization to the outputs of activation functions during training. Left: DMC rewards. Right: MW success. Applying SN to the outputs of both Original Rationals (OR) and Constrained Rationals (CR) hinders training, often leading to degraded or stalled performance. These results contrast with the effectiveness of SN when applied to weight matrices (Figure~\ref{fig:sn-rl-results}), indicating that output normalization suppresses useful functional dynamics in learnable activations.}
    \label{fig:dmc-mw-activations-sn-results}
\end{figure}

Figure~\ref{fig:dmc-mw-activations-sn-results} presents the results of this intervention. We observe that, contrary to expectations, applying SN directly to activation outputs degrades performance across both DMC and MW benchmarks. In particular, for CR+SN and OR+SN, learning is either slowed or entirely disrupted, especially in the early training phases. Compared to baseline CR and OR, the application of SN on activations fails to preserve the performance benefits typically associated with these function classes.

These results suggest that output normalization interferes with the learning dynamics of expressive activation functions, particularly when these functions rely on learning nontrivial output distributions. In essence, normalization suppresses the very flexibility that makes rational functions effective, masking rather than resolving pathological behaviors. This aligns with our broader hypothesis: the efficacy of learnable activations hinges not just on controlling magnitude, but on preserving functional plasticity during training.

\begin{figure}[h!]
    \centering
    \parbox{\textwidth}{ % Umieszcza pierwszy obrazek w osobnym wierszu
        \centering
        \includegraphics[width=0.6\linewidth]{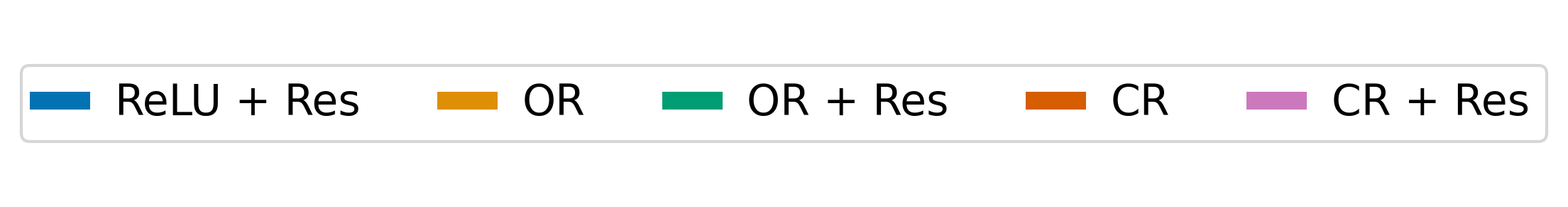}
    }
    \includegraphics[width=0.19\linewidth]{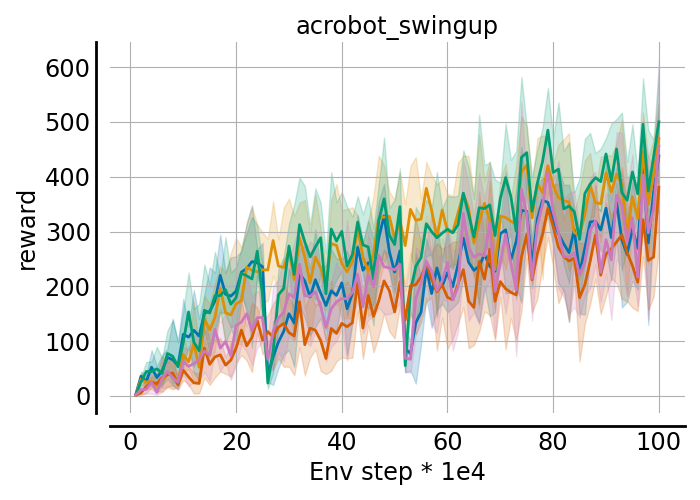}
    \includegraphics[width=0.19\linewidth]{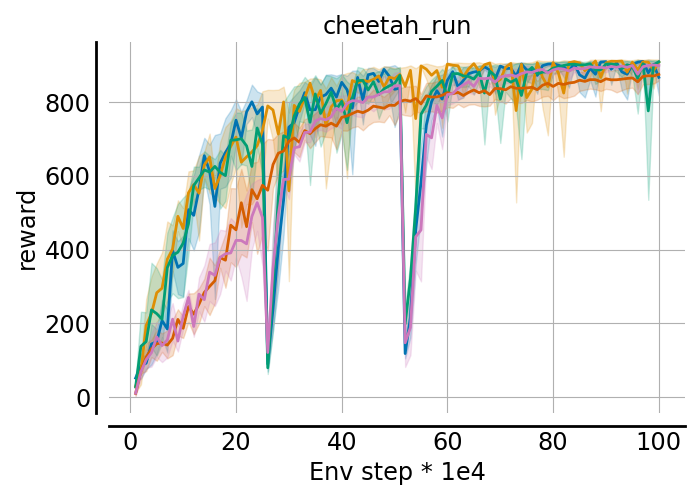}
    \includegraphics[width=0.19\linewidth]{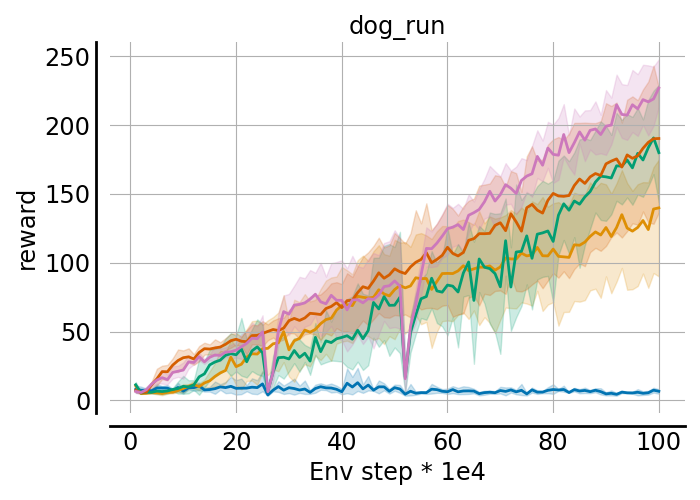}
    \includegraphics[width=0.19\linewidth]{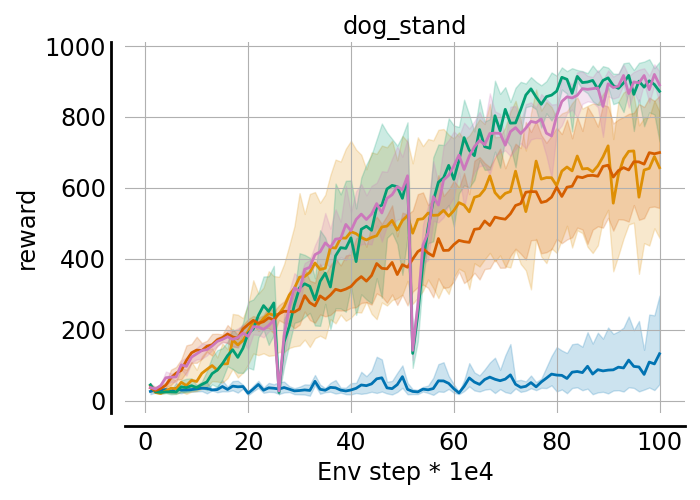}
    \includegraphics[width=0.19\linewidth]{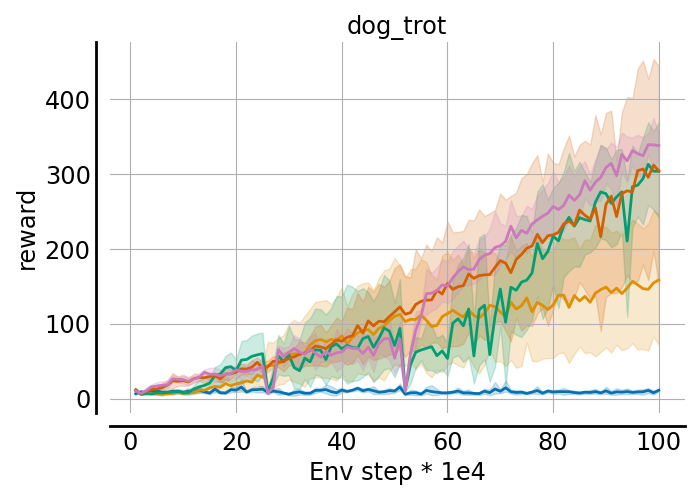}
    \includegraphics[width=0.19\linewidth]{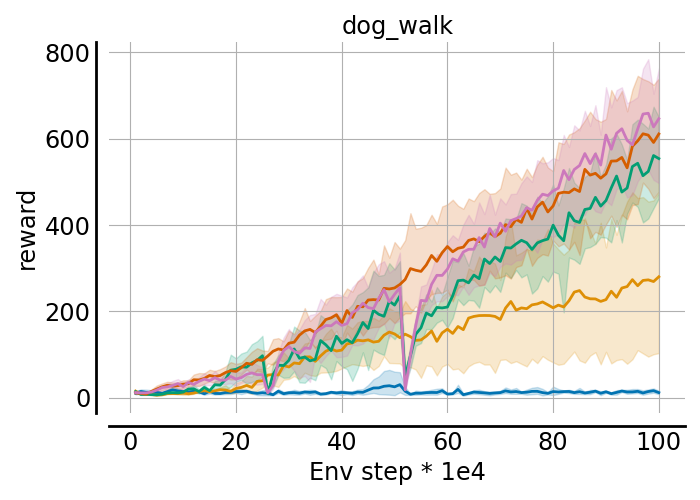}
    \includegraphics[width=0.19\linewidth]{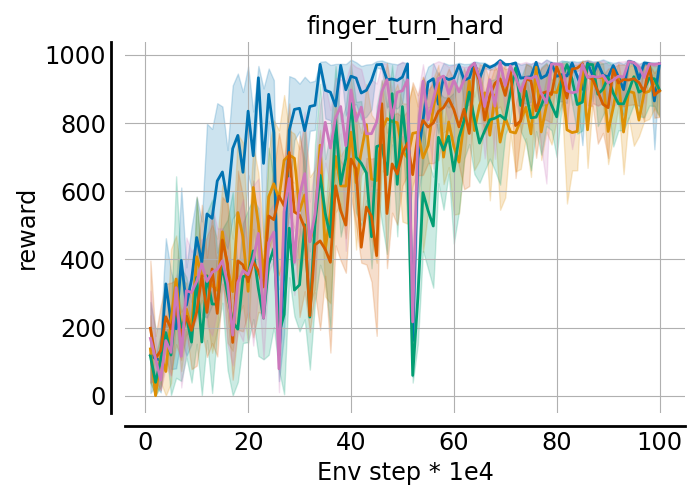}
    \includegraphics[width=0.19\linewidth]{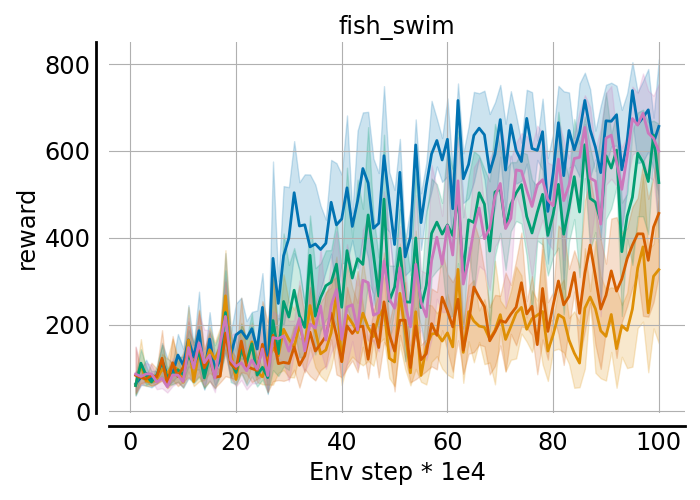}
    \includegraphics[width=0.19\linewidth]{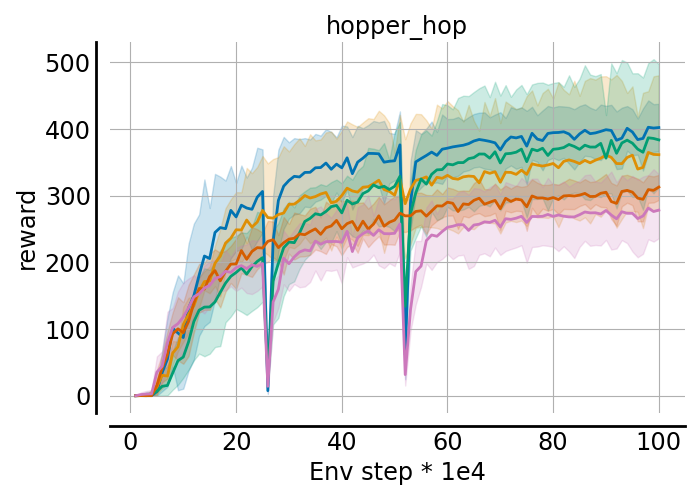}
    \includegraphics[width=0.19\linewidth]{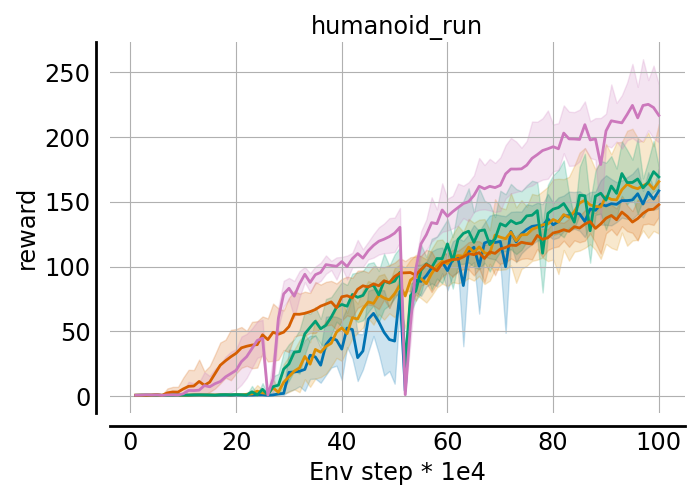}
    \includegraphics[width=0.19\linewidth]{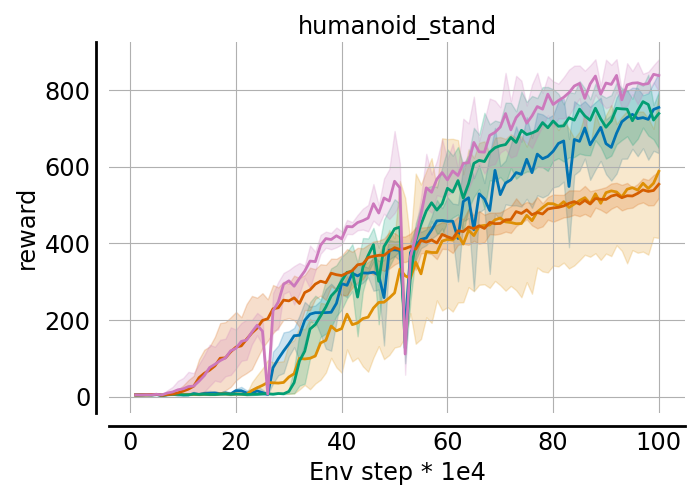}
    \includegraphics[width=0.19\linewidth]{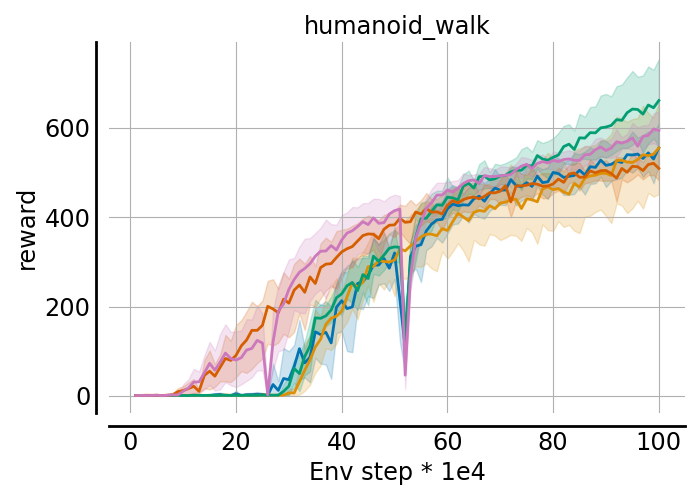}
    \includegraphics[width=0.19\linewidth]{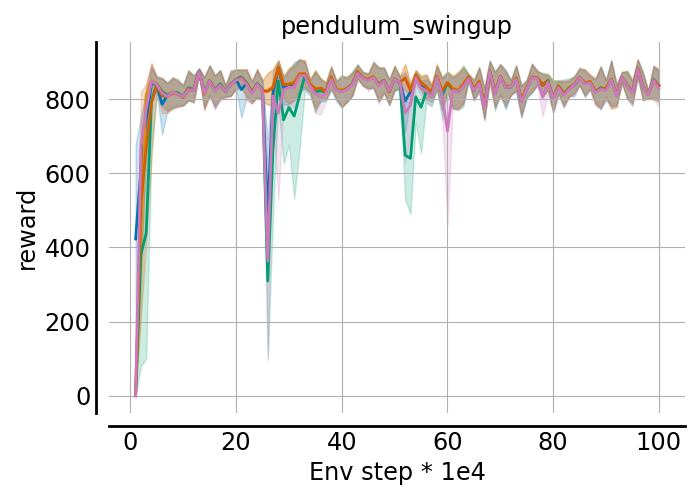}
    \includegraphics[width=0.19\linewidth]{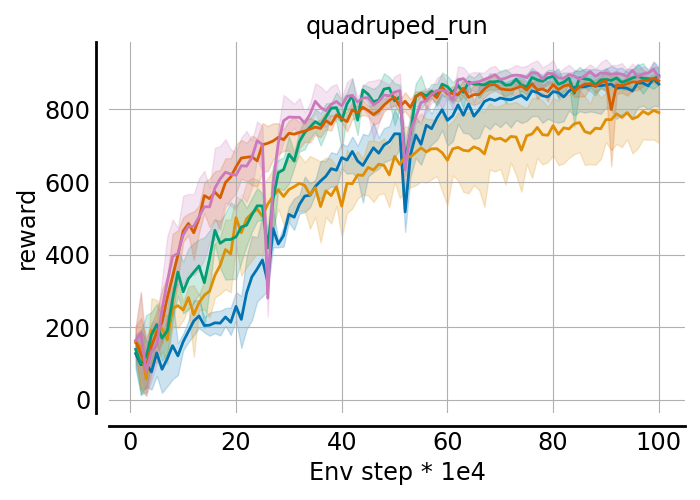}
    \includegraphics[width=0.19\linewidth]{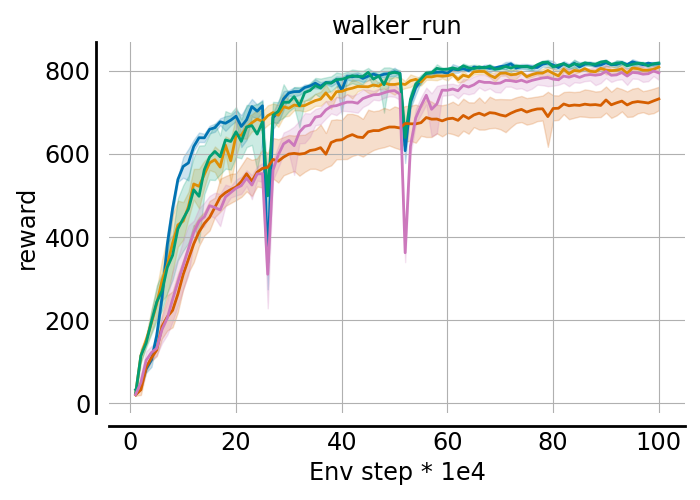}
    \caption{Results of 15 tasks from \textbf{DeepMind Control Suite} for different setups for 1M steps. We present the mean of rewards and 95\% confidence intervals.}
    \label{fig:dmc-envs-performance}
\end{figure}

\begin{figure}[h!]
    \centering
    \parbox{\textwidth}{ % Umieszcza pierwszy obrazek w osobnym wierszu
        \centering
        \includegraphics[width=0.6\linewidth]{figs/envs/legend4.png}
    }
    \includegraphics[width=0.19\linewidth]{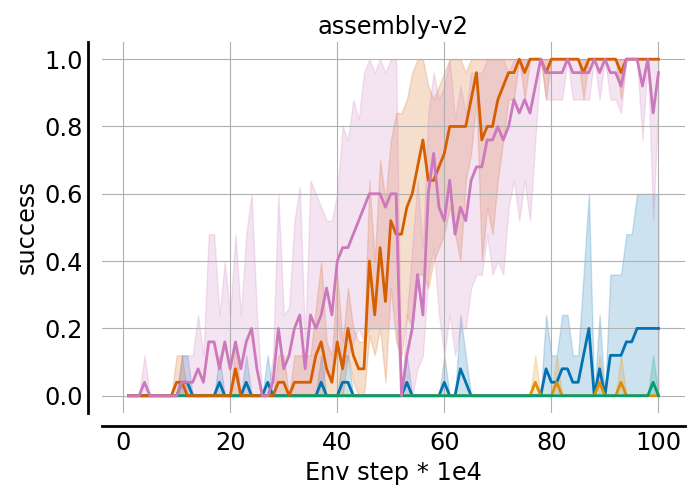}
    \includegraphics[width=0.19\linewidth]{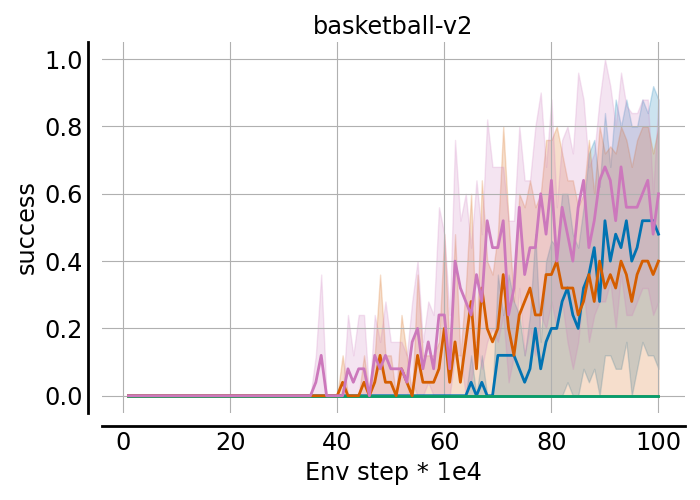}
    \includegraphics[width=0.19\linewidth]{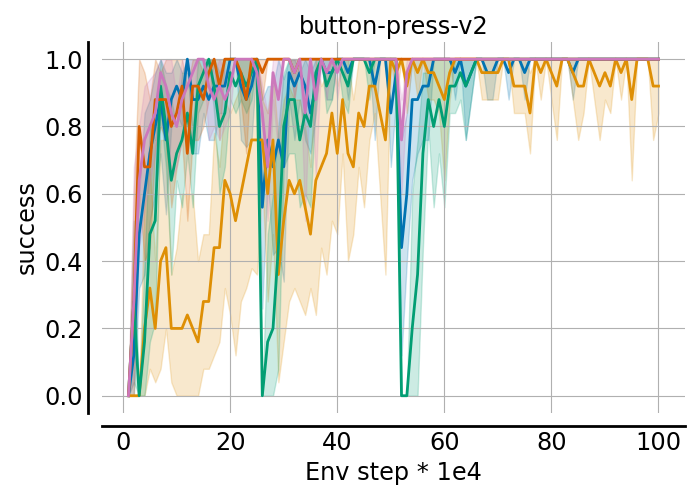}
    \includegraphics[width=0.19\linewidth]{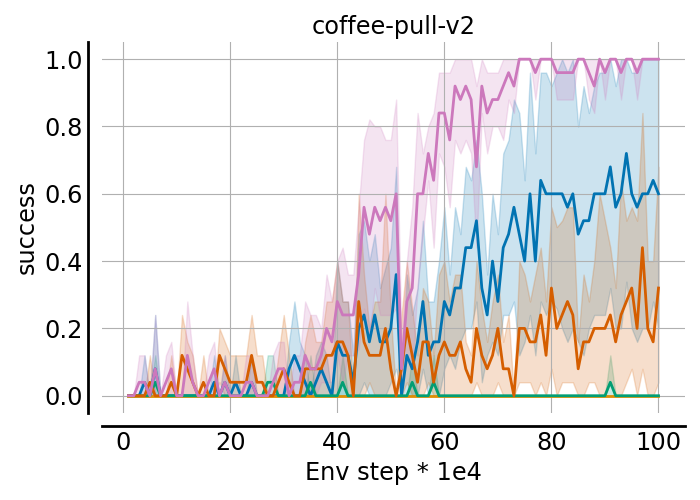}
    \includegraphics[width=0.19\linewidth]{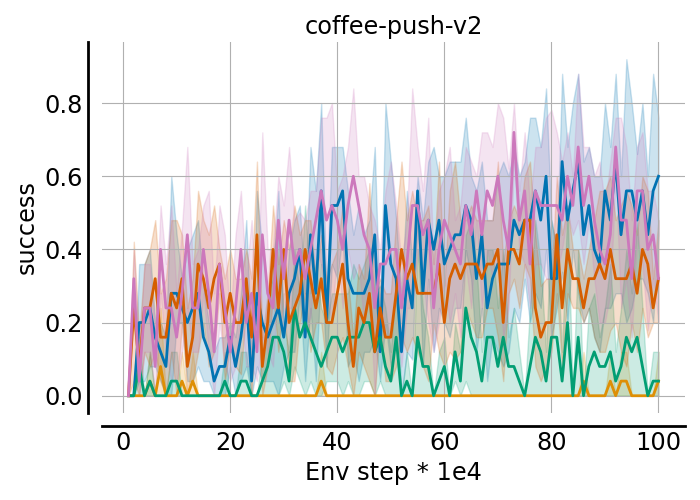}
    \includegraphics[width=0.19\linewidth]{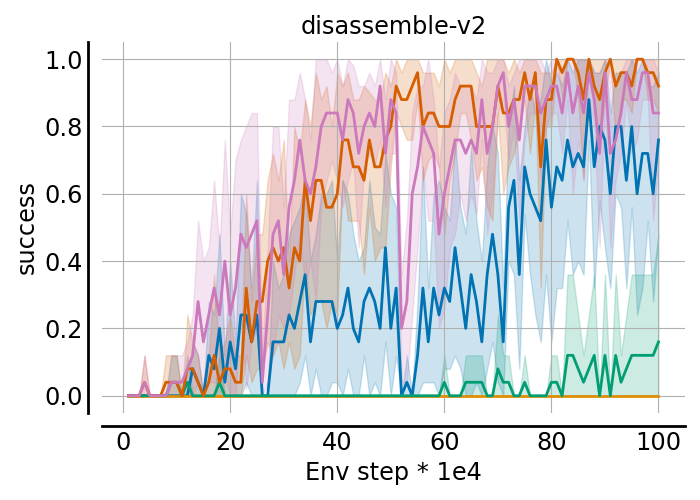}
    \includegraphics[width=0.19\linewidth]{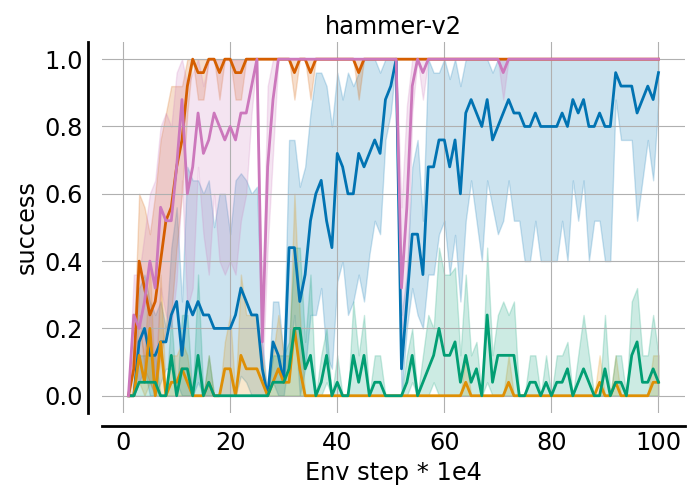}
    \includegraphics[width=0.19\linewidth]{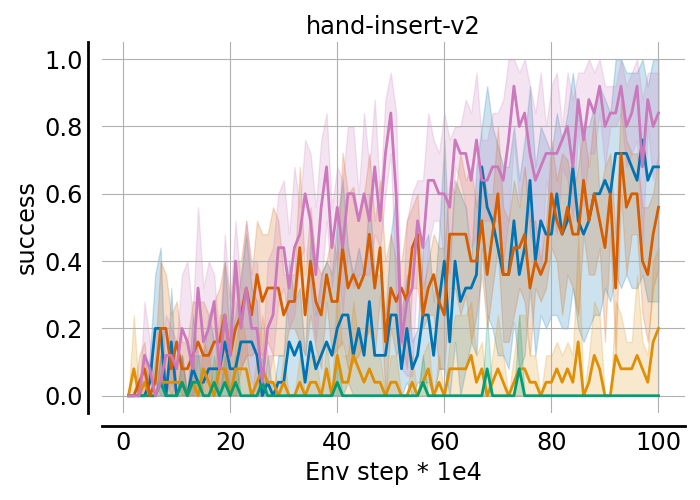}
    \includegraphics[width=0.19\linewidth]{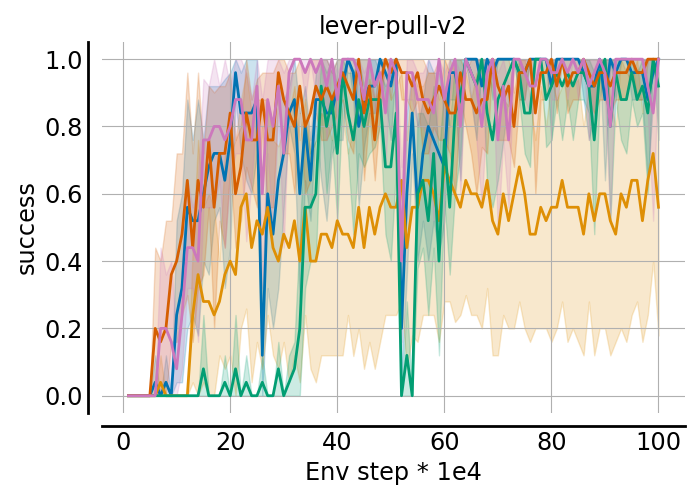}
    \includegraphics[width=0.19\linewidth]{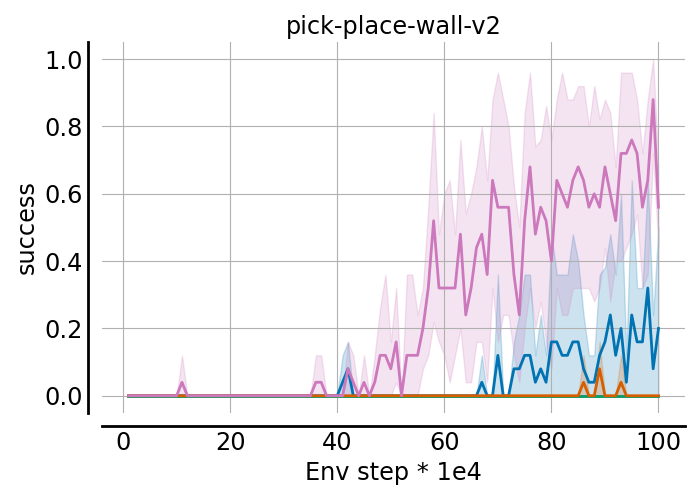}
    \includegraphics[width=0.19\linewidth]{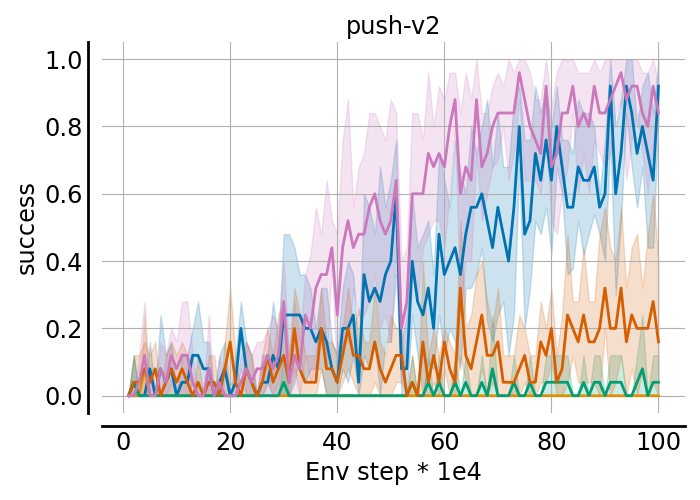}
    \includegraphics[width=0.19\linewidth]{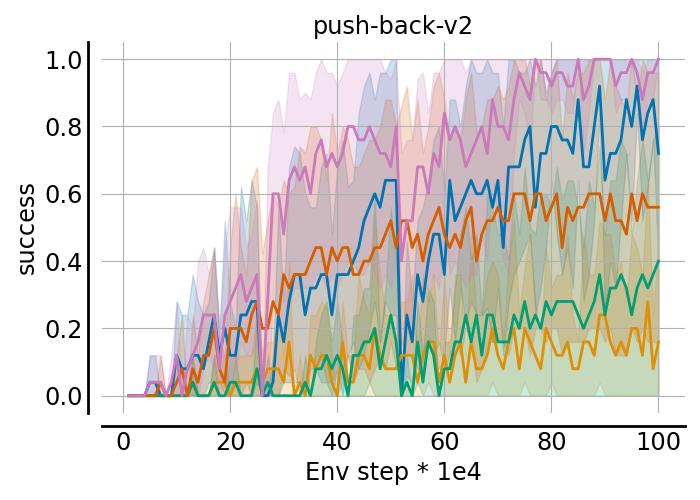}
    \includegraphics[width=0.19\linewidth]{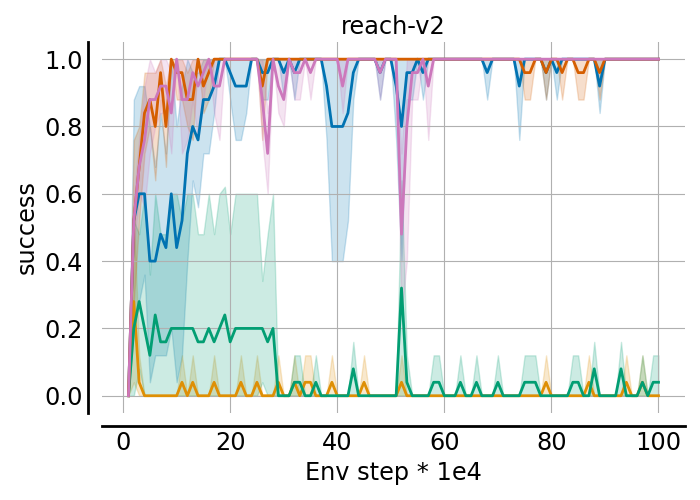}
    \includegraphics[width=0.19\linewidth]{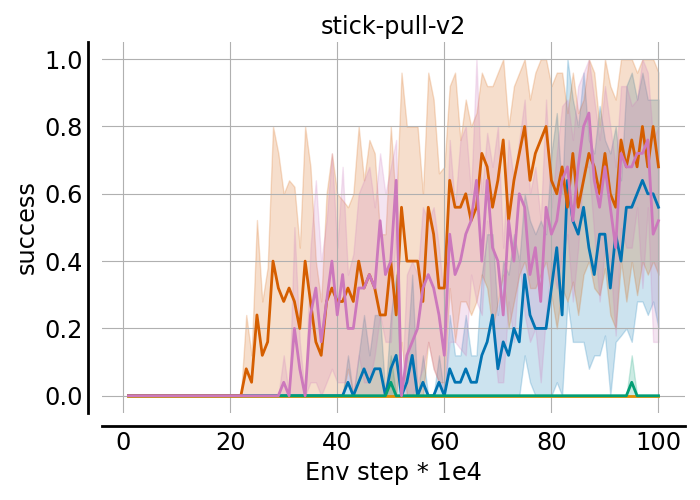}
    \includegraphics[width=0.19\linewidth]{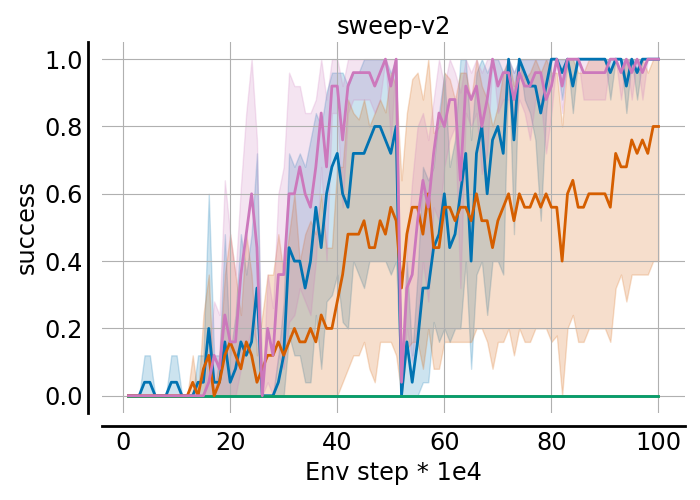}
    \caption{Results of 15 tasks from \textbf{MetaWorld} for different setups for 1M steps. We present the mean of success rate and 95\% confidence intervals.}
    \label{fig:mw-envs-performance}
\end{figure}

\subsection{NTK analysis in reinforcement learning}
\label{appendix:rl-ntk-heatmaps}
\begin{figure}
    \includegraphics[width=\linewidth]{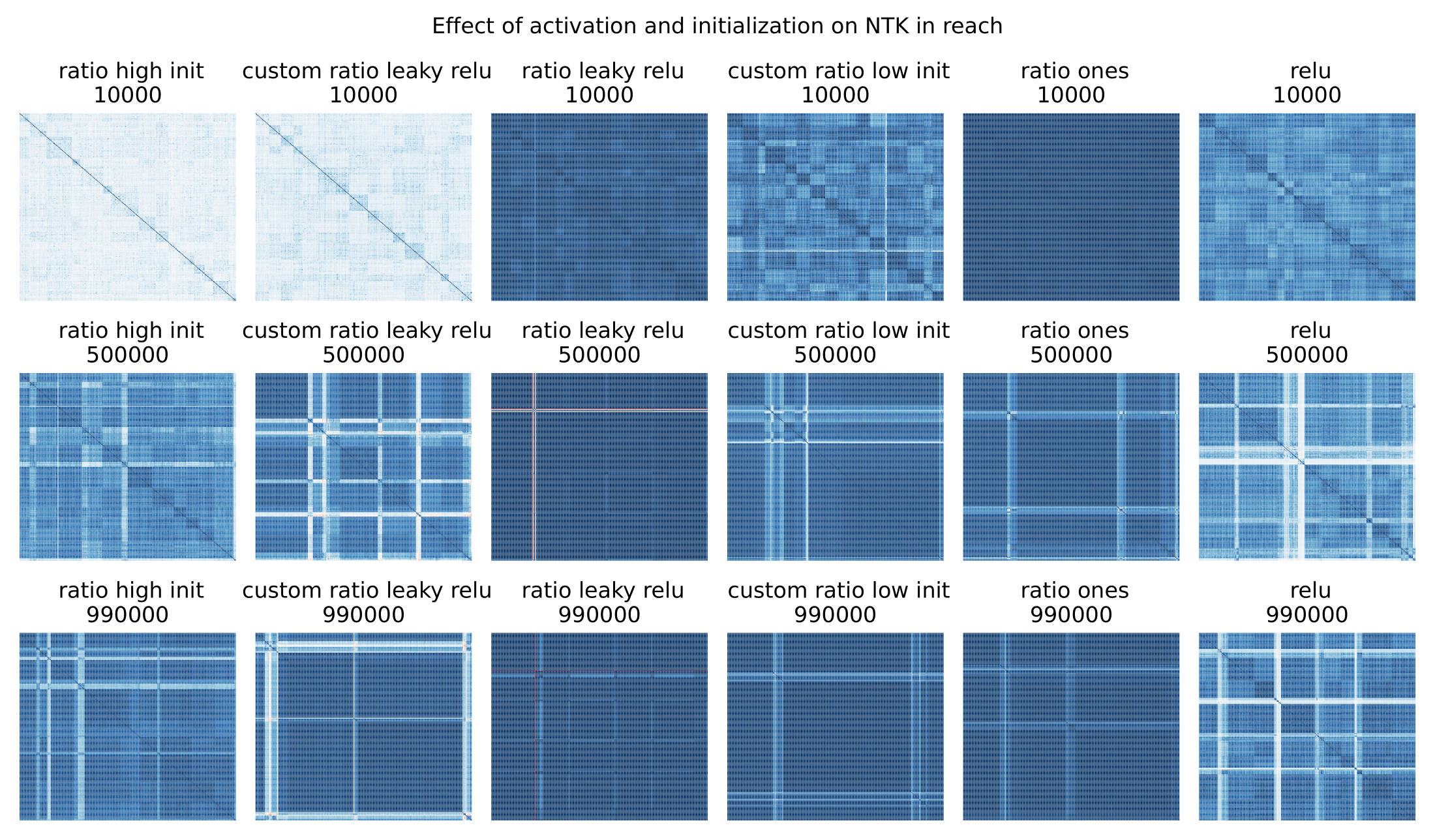}
    \caption{Visualization of empirical NTK computed on a random sampling of 256 observations. Each row corresponds to a different timestep during training, and each column corresponds to a different initialization and parameterization method. All samples are drawn for seed 0.}
    \label{fig:ntk-heatmap-1}
\end{figure}

\begin{figure}
    \includegraphics[width=\linewidth]{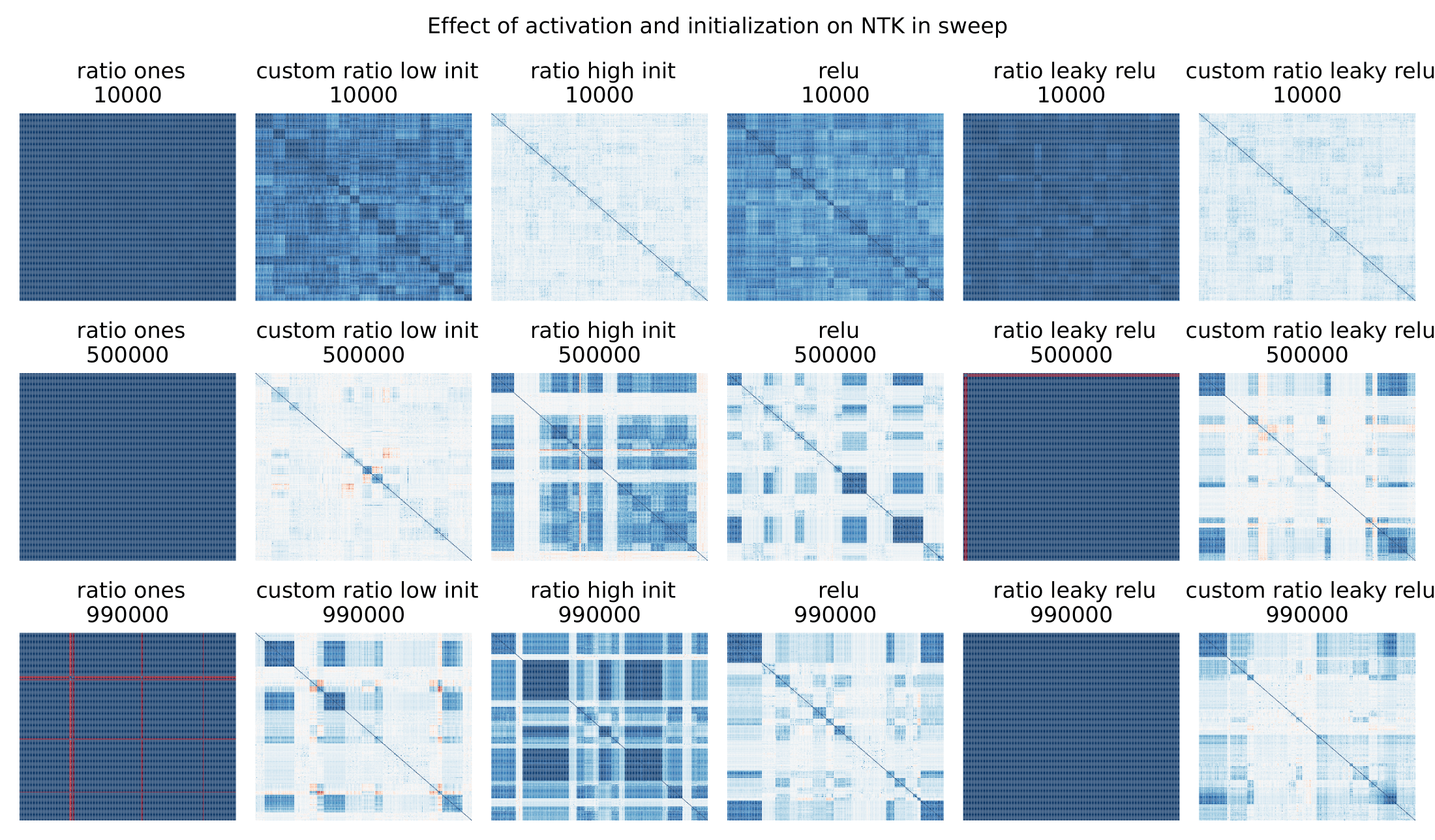}
    \caption{Visualization of empirical NTK computed on a random sampling of 256 observations. Each row corresponds to a different timestep during training, and each column corresponds to a different initialization and parameterization method. All samples are drawn for seed 0.}
    \label{fig:ntk-heatmap-2}
\end{figure}

In this section, we provide additional details on the empirical computation of NTK mentioned in Section~\ref{sec:initialization}. The NTK matrices were computed every 10k environment steps for a batch of observations and actions of size 256. Figures~\ref{fig:ntk-heatmap-1} and \ref{fig:ntk-heatmap-2} visualize the NTK for different activation functions with various initializations. The visualizations are presented for steps 10k (early training), 500k (mid-training), and 990k (late training).

\FloatBarrier

\label{appendix:atari-results}
\section{Reinforcement learning on Atari 100k}

To investigate whether the instability issues associated with rational activation functions also manifest in discrete-action environments, we conducted a small-scale study using five Atari 100k games from the Atari-5 benchmark: \textit{BattleZone}, \textit{Jamesbond}, \textit{NameThisGame}, \textit{SpaceInvaders}, and \textit{Qbert}. We used the SPR~\citep{schwarzer2021dataefficientreinforcementlearningselfpredictive} algorithm as our base method and compared three types of activation functions: standard ReLU, Original Rational (OR), and Constrained Rational (CR). For each variant, we also evaluated the effect of periodically resetting all trainable parameters in the agent (denoted as “+Resets”).

\begin{figure}
    \centering
    \includegraphics[width=0.5\linewidth]{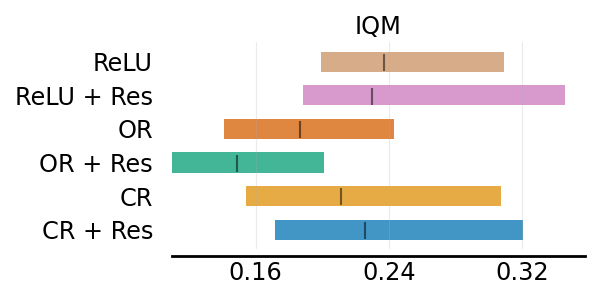}
    \includegraphics[width=0.5\linewidth]{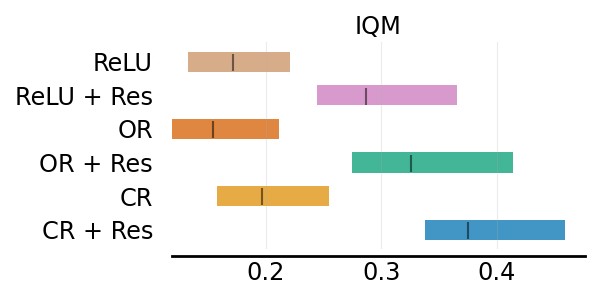}
    \includegraphics[width=0.5\linewidth]{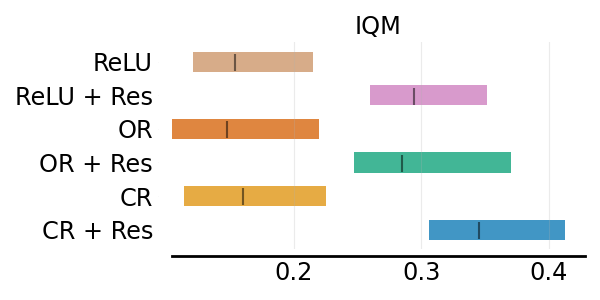}
    \caption{IQM performance across five Atari 100k games (\textit{BattleZone}, \textit{Jamesbond}, \textit{NameThisGame}, \textit{SpaceInvaders}, \textit{Qbert}) over 20 random seeds for different activation function variants (ReLU, Original Rational (OR), Constrained Rational (CR)) with and without coefficient resets. Top: UTD = 1, Middle: UTD = 16, Bottom: UTD = 32. Resets consistently help across all UTD settings, and CR+Resets shows the highest average performance, particularly as UTD increases, though results are not statistically conclusive due to wide confidence intervals.}
    \label{fig:Atari_iqm}
\end{figure}

\paragraph{Results}

Figure~\ref{fig:Atari_iqm} shows the Interquartile Mean (IQM) performance aggregated over 20 random seeds. While the sample size is limited, consistent patterns emerge across runs, suggesting that the core stability issues observed in continuous control do not generalize to discrete-action domains like Atari.

\begin{itemize} 
\item Across all activation types, applying periodic coefficient resets consistently improves performance, supporting earlier findings from our continual learning and control experiments.
\item The CR+Resets variant yields the highest average performance, particularly in high replay ratio settings, suggesting that constraining the activation functions while periodically reinitializing them can boost stability and learning efficiency. 
\item Unlike in continuous control environments, we did not observe severe instability or divergence when using rational activations in Atari.
\end{itemize}

\paragraph{Gradient Dynamics}

\begin{figure}
    \centering
    \includegraphics[width=0.45\linewidth]{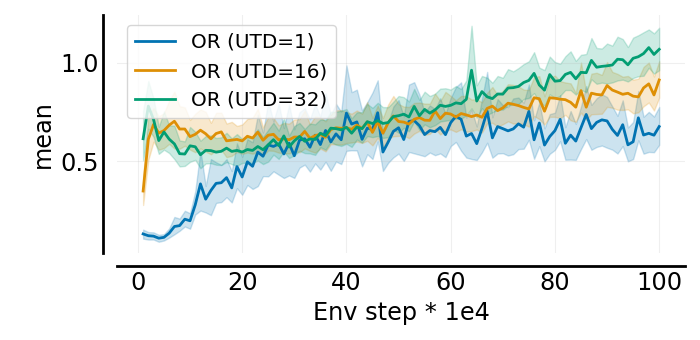}
    \includegraphics[width=0.45\linewidth]{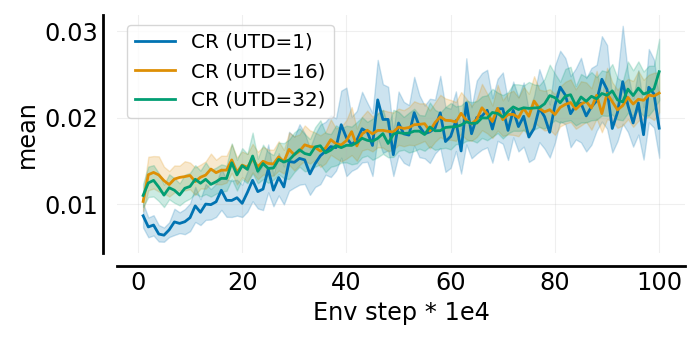}
    \caption{Gradient norms of the numerator and denominator coefficients of Original Rational (OR) activations during training on Atari 100k across different UTD settings (UTD = 1, 16, 32). Unlike in continuous control environments, gradient norms remain bounded and stable throughout training, indicating no signs of coefficient explosion even at higher replay ratios.}
    \label{fig:Atari_gnorm}
\end{figure}

To better understand why rational activations behave more stably in discrete environments, we examined the gradient norms of the trainable coefficients during training. As shown in Figure~\ref{fig:Atari_gnorm}, gradient magnitudes remain small throughout training, indicating that coefficient updates do not grow uncontrollably, in contrast to what we observed in continuous control settings. On the other hand, we observe that the gradient norm increases with higher UTD values for Original Rationals, which may indicate a potential risk of parameter instability under certain conditions. For the Constrained Rationals, however, gradient norms remain stable across different UTD settings, further substantiating its robustness.

\paragraph{Conclusion}

These findings suggest that rational activations are more stable in discrete-action environments like Atari, likely due to less dynamic input distributions compared to continuous control tasks. While further experiments with more environments are needed for stronger conclusions, the observed trends support our hypothesis that the instability of rational activations is context-dependent.
\section{Continual Learning Experiments}
\label{appendix:cl-experiments}

In this section, we provide a detailed analysis of the impact of coefficient initialization on plasticity in continual learning, based on experiments conducted with the MNIST with non-stationary targets and Split/Cifar-100 benchmarks.

\subsection{Adaptability on Split/Cifar-100}
\label{appendix:split-cifar}
\begin{figure}
    \centering
    \includegraphics[width=0.5\linewidth]{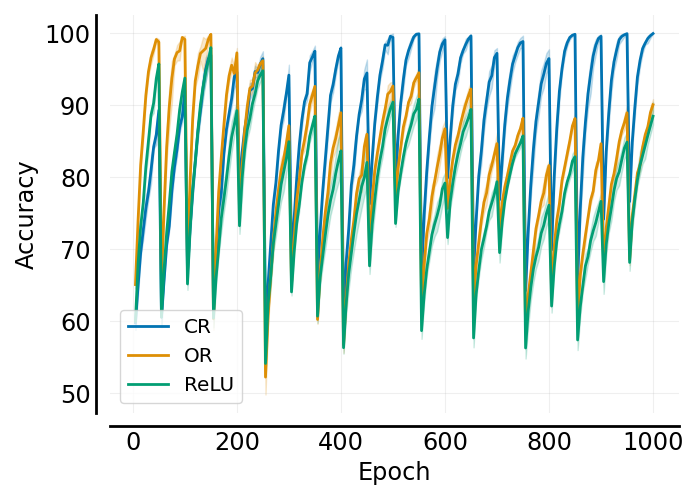}
    \caption{Training accuracy for the Split/CIFAR-100 experiment. Comparison of rational activation functions with ReLU. Our Constrained Rationals demonstrate improved adaptability in later tasks.}
    \label{fig:cifar}
\end{figure}

In addition to the MNIST with shuffled labels experiment, we also conducted a continual learning experiment on the CIFAR-100 dataset. The data set was divided into 20 subsets, each containing five classes. The experiment involved training an MLP with two hidden layers for 50 epochs on each subset. Each layer consisted of 512 hidden neurons, and the activation functions differed for each setup. The results are presented in Figure~\ref{fig:cifar}.

It can be seen that our Constrained Rationals achieve superior accuracy compared to networks using ReLU and Original Rationals especially in the final tasks. This is consistent with our conclusions from Section~\ref{sec:dynamics}, where we mentioned that our constrained activation functions make better use of individual samples.

However, these results should be interpreted with caution in the context of network plasticity. It is important to note that each task (subset) in Split/CIFAR-100 contained significantly fewer training examples, which the network had to fit to, and the number of tasks was three times lower compared to MNIST experiment.

\subsection{Effect of Coefficient Initialization on Plasticity}

\begin{figure}[h!] \centering \includegraphics[width=0.45\linewidth]{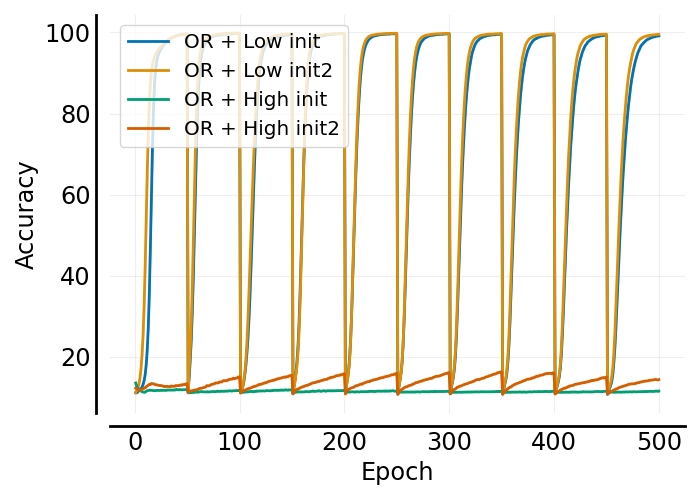} \includegraphics[width=0.45\linewidth]{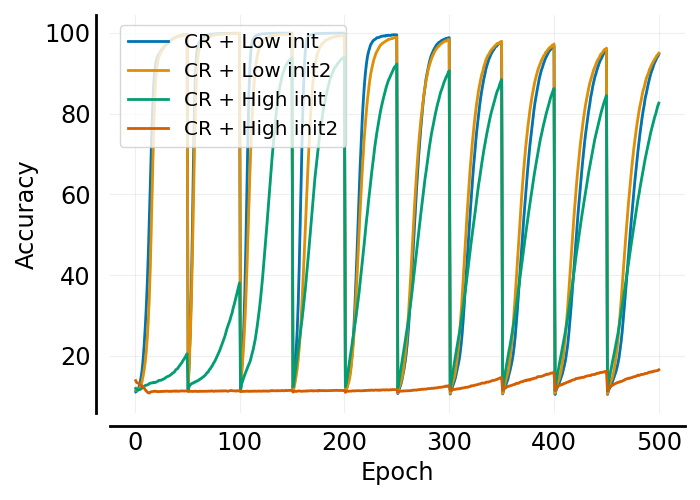} \caption{MNIST with non-stationary targets: Comparison of different coefficient initialization strategies. Functions initialized with smaller coefficients retain plasticity longer.} \label{fig:pmnist-inits} \end{figure}

To examine the role of initialization, we tested two coefficient initialization schemes: one with small values and another with larger values. The results, shown in Figure~\ref{fig:pmnist-inits}, reveal a clear pattern: networks initialized with smaller coefficients maintain significantly higher plasticity throughout training, enabling them to adapt to new tasks more effectively.

\subsection{NTK Analysis in Continual Learning}

To better understand these findings, we analyze the Neural Tangent Kernel (NTK) spectrum for different initialization schemes. Our results reveal a striking trend:

\begin{itemize} \item \textbf{Constrained Rationals} maintain the maximum NTK rank (256) regardless of initialization.
\item \textbf{Original Rationals} initialized with high coefficients also achieve the maximum NTK rank, whereas those initialized with low coefficients exhibit a reduced rank of 235.
\end{itemize}

These patterns align with our empirical observations: Original Rationals with low initialization effectively model the relatively simple MNIST dataset and maintain plasticity over multiple tasks. The lower NTK rank in this case suggests that the function space remains more constrained, preserving adaptability. In contrast, a higher-rank NTK, while offering greater expressivity, appears to accelerate feature collapse, reducing the network's ability to retain knowledge over time.

\end{document}